\documentclass[sigconf, nonacm]{acmart}

\usepackage{hyperref}
\usepackage{url}
\usepackage{graphicx}
\usepackage{booktabs}       %
\usepackage{amsfonts}       %
\usepackage{nicefrac}       %
\usepackage{multirow}
\usepackage{mathrsfs}
\usepackage{wrapfig}
\usepackage{color}
\usepackage{algorithm}
\usepackage{algpseudocode}
\usepackage{setspace}
\usepackage{marvosym}
\usepackage{subcaption}
\usepackage{tabularx}
\usepackage{bbm}
\usepackage{caption}
\usepackage{listings}  %

\newcommand\vldbpagestyle{plain}

\begin{document}
\title{AMAD: AutoMasked Attention for  Unsupervised Multivariate Time Series Anomaly Detection}

\author{Tiange Huang}
\affiliation{%
  \institution{Northwestern Polytechnical University}
  \city{Xi'an}
  \state{China}
}
\email{kevinhuang@mail.nwpu.edu.cn}

\author{Yongjun Li*}
\affiliation{%
  \institution{Northwestern Polytechnical University}
  \city{Xi'an}
  \state{China}
}
\email{lyj@nwpu.edu.cn}

\begin{abstract}

Unsupervised multivariate time series anomaly detection (UMTSAD) plays a critical role in various domains, including finance, networks, and sensor systems. In recent years, due to the outstanding performance of deep learning in general sequential tasks, many models have been specialized for deep UMTSAD tasks and have achieved impressive results, particularly those based on the Transformer and self-attention mechanisms. However, the sequence anomaly association assumptions underlying these models are often limited to specific predefined patterns and scenarios, such as concentrated or peak anomaly patterns. These limitations hinder their ability to generalize to diverse anomaly situations, especially where the lack of labels poses significant challenges. 

To address these issues, we propose AMAD, which integrates \textbf{A}uto\textbf{M}asked Attention for UMTS\textbf{AD} scenarios. AMAD introduces a novel structure based on the AutoMask mechanism and an attention mixup module, forming a simple yet generalized anomaly association representation framework. This framework is further enhanced by a Max-Min training strategy and a Local-Global contrastive learning approach. By combining multi-scale feature extraction with automatic relative association modeling, AMAD provides a robust and adaptable solution to UMTSAD challenges.

Extensive experimental results demonstrate that the proposed model achieving competitive performance results compared to SOTA benchmarks across a variety of datasets.

\end{abstract}

\maketitle

\pagestyle{\vldbpagestyle}
\begingroup
\renewcommand\thefootnote{}\footnote{\noindent

* corresponding author.

}\addtocounter{footnote}{-1}\endgroup

\section{Introduction}
Anomaly detection in multivariate time series data is a critical task in various domains and systems such as finance, network systems, sensor networks, and industrial automation. These systems generate vast amounts of time series data, where timely and accurate anomaly detection is essential for maintaining operational efficiency, security, and reliability. However, the lack of labeled data and the complex, high-dimensional nature of multivariate time series pose significant challenges for traditional anomaly detection methods.\cite{li2019madganma,su2019ominianomaly,tuli2022tranad,Yairi2017ADH,zhang2021caem,chatfield1978holtwinters,shumway2017arima,le2022log}

\begin{figure}[ht]
  \centering

  \includegraphics[width=0.5\textwidth]{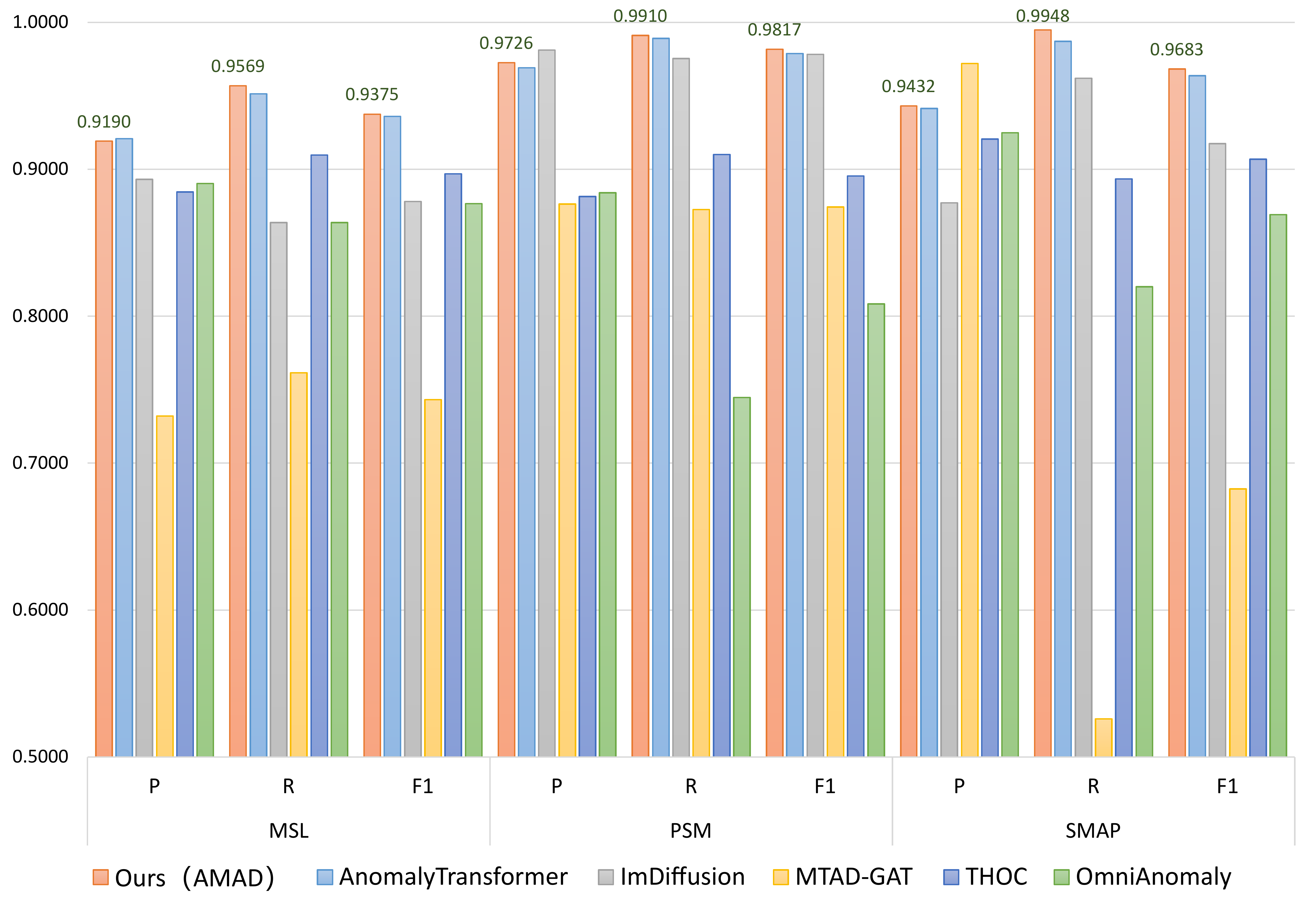}

  \caption{
  The experimental results of our method (orange bars) compared to five SOTA methods on three datasets. The results show that our method outperforms the others in most metrics.}
  \label{fig:toc}
\end{figure}

Given that real-world systems accumulate vast amounts of data, with anomalies constituting only a minor and non-uniformly distributed portion, labeling such data is both costly and impractical. Consequently, the focus of our research is on unsupervised or self-supervised time series anomaly detection. %

Unsupervised time series anomaly detection presents substantial practical challenges. Due to the 
lack of labeled data, training cannot rely on labels but must instead depend on well-designed unsupervised or self-supervised tasks to train sequence representation models. These representation models must be sufficiently robust to learn discriminative features that can distinguish various characteristics across different sequential data. Consequently, they should be capable of generating feature scores with significant discriminatory, enabling the identification of a small number of anomalies that deviate from normal sequences within large volumes of time series data. The success of this approach hinges on the model's ability to capture complex patterns and nuances inherent in the data without explicit guidance from labeled examples, thereby effectively isolating anomalies based purely on learned representations of normal behaviors.

\subsection{Related Works}

Traditional approaches to UMTSAD often rely on simplistic assumptions or are constrained to specific types of anomalies, limiting their ability to represent generalized anomaly patterns. These methods typically utilize only a limited amount of information from the sequences, focusing on individual time points rather than capturing the broader context of the data. For instance, models based on the Transformer architecture calculate attention based on the self-similarity of each point within the sequence, thereby restricting their focus to local point information. This limitation hinders their ability to effectively detect anomalies that are characterized by more complex temporal dependencies and interactions.\cite{zhao2020mtadgat,zhou2019beatganar,malhotra2015lstmad}%

THOC\cite{THOC} incorporates multi-scale features of sequences into its model structure through the use of dilated RNNs. However, due to the assumption of hypersphere optimization, THOC can also be categorized as a special type of association model concerning the hypersphere. Furthermore, the association model of THOC is based on the relationships between data points in the sequence rather than local associations. Meanwhile, the stacked RNN architecture increases the paths for mutual learning between sequence data points. According to studies on sequential models\cite{transformer,hochreiter2001transpath}, shorter paths facilitate better learning of sequence representations by the model. Therefore, theoretically, the association model of THOC faces similar issues.

Models utilizing the Transformer architecture theoretically encounter similar limitations, as attention is computed based on the self-similarity between each point and the sequence, thereby being confined to point-wise information. To address this issue, AnomalyTransformer \cite{xu2021anomalytrans}introduces local information through associated differences to distinguish anomalies from normal data points. This approach relies on a Gaussian kernel prior assumption that the differentiation of anomaly points is negatively correlated with the distance to local points; Chronos\cite{ansari2024chronos} draws inspiration from masked language models akin to BERT\cite{devlin2019bert}, employing random masking as a source for the model to learn dependencies within the sequence; ImDiffusion\cite{chen2023imdiffusion} adopts a denoising diffusion model to interpolate a series of states in the sequence, thus implicitly modeling the degree of association within the sequence.\cite{transformer,zhou2021informer,wu2021autoformer}

Existing Transformer-based methods often neglect to consider multi-scale anomaly correlations, meaning they do not simultaneously account for both local and global associations. To this end, we propose a variant of the Transformer model designed to capture multi-scale anomaly correlation structures. This model improves upon the self-attention mechanism, which is limited to computing only global association features, by introducing a novel mechanism called \emph{AutoMaskAttention}.

\subsection{Our Contribution}

Real-world network systems collect vast amounts of traffic data, with anomalies constituting only a small and non-uniformly distributed portion. Labeling such data is both expensive and impractical; therefore, the current focus of research is on unsupervised time series anomaly detection.
This task is also highly challenging in practice:

On one hand, due to the nature of unsupervised learning, the representation model must be sufficiently generalized to learn distinguishing feature representations across various types of sequence data. This allows the model to produce feature scores with significant discriminative power.

On the other hand, without labels for aligning model parameters with the task during training, the model relies solely on well-designed self-supervised tasks to learn sequence representations. The goal is to detect a small number of anomalies that deviate from normal sequences within large volumes of time series data. These self-supervised tasks must converge to an appropriate local optimum during training to support subsequent anomaly detection tasks. Defining reasonable self-supervised tasks is thus a critical challenge.

To address these limitations, we propose AMAD (Auto Masked Attention for Unsupervised Multivariate Time Series Anomaly Detection), a novel framework designed to enhance the detection of anomalies by capturing both local and global features of time series data. AMAD introduces a more generalized association perspective, enabling the characterization of sequence correlations across various distances. This is achieved through the introduction of an associative function, \(Association(x_i, x_j) = K(x_i, x_j, i - j)\), where \(K\) is a generalized distance function of the sequence, thereby modeling sequence correlations in a comprehensive manner.

AMAD leverages the theoretical foundation of Fourier Transformation, which suggests that a function can be approximated by a linear combination of periodic functions. By analogy, we extend the concept of Rotary Position Encoding (RoPE) 
\cite{su2024roformer} to approximate arbitrary functions, similar to the basis functions in Fourier series. This auto mask mechanism allows the model to learn representations that can distinguish between normal and anomalous patterns by capturing the intricate relationships within the data.
Our approach ensures that the model can effectively capture the temporal dynamics of the sequences, moving beyond the constraints of point-based analysis.

Furthermore, AMAD incorporates a mixup technique, which acts as a gate facility to regulate the flow of information within the model. This feature enhances the model's ability to discern between normal and anomalous patterns by blending information from different data points in a controlled manner. The Max-Min training strategy employed by AMAD prevents the model from descending into a trivial solution and enhances its sensitivity to less significant anomalies.

In summary, AMAD represents an advancement in the field of UMTSAD. By combining multiscale feature extraction and automatic rotary mask, AMAD provides a robust and adaptable solution to the challenges of detecting anomalies in real-world time series data. This paper details the design and implementation of AMAD and presents experimental results that demonstrate its effectiveness compared to existing methods.

The main contributions of our work are listed as follows:

1) \emph{AMAD}: Our proposed model addresses the shortcomings of existing Transformer-based methods by considering multi-scale anomaly correlations.

2) \emph{AutoMask Attention}: A novel attention mechanism integrates multi-scale sequence relative position information to capture both local and global sequence correlations.

3) \emph{Attention Mixup}: A straightforward Mixup module which is used to fuse local relative sequence information with overall feature information, ensuring comprehensive feature representation.
    
4) \emph{Self-Supervised Strategy}: The attention modules are optimized using Max-Min strategy and Local-Global contrastive strategy, and the model employs a reconstruction task to construct a global loss function, facilitating robust training and evaluation.

\section{Preliminary}

In this section, we initially present a formal definition of sequential data and problem about UMTSAD. Subsequently, we introduce the definition of anomaly correlations.

\begin{table}[htb]
\label{tab:symbol_table}
\centering
\caption{Symbols Used in the Paper}
\begin{tabular}{c p{6.7cm}}
\toprule
\textbf{Symbol} & \textbf{Description} \\ 
\midrule
$d$ & The dimension of the system state, indicating the number of variables observed. \\ 
$N$ & The length of the time series, indicating the total number of observations. \\
$t$ & Discrete time index, ranging from $1$ to $N$. \\ 
$\mathbf{x}$ & A $d$-dimensional vector representing the system state in $\mathbb{R}^{d}$. \\ 
$\mathcal{T}$ & A time series: an ordered sequence of system states $\{\mathbf{x}_1, \mathbf{x}_2, \ldots, \mathbf{x}_N\}$. \\ 
$f(\mathcal{T})$ & A time series model that takes $\mathcal{T}$ as input and outputs a binary vector $\mathbf{y}$. \\ 
$\mathbf{y}$ & A binary vector $\mathbf{y} = [y_1, y_2, \ldots, y_N]$, where $y_t \in \{0, 1\}$ denotes normal ($0$) or anomalous ($1$). \\ 
$\mathbbm{1}_{threshold}$ & The anomaly detection result by threshold used to classify states as normal ($\mathbbm{0}$) or anomalous ($\mathbbm{1}$). \\ 
\bottomrule
\end{tabular}
\label{tab:symbols}
\end{table}

\subsection{Time Series Definition}

In the following, we present the fundamental definitions of system state and time series.

\emph{Definition 1 (System State)}
The \emph{system state} $\mathbf{x}$ is a $d$ dimensional vector in $ \mathbb{R}^d $,
where each element $x_{i}$ represents the value of the $i$-th state variable observed at time $t$.

\emph{Definition 2 (Time Series)} 
A \emph{time series} is a sequence of data points, defined as $\mathcal{T}$, ordered by the time $t \in \{1, \ldots , N\}$ at which they were observed:
$$
\mathcal{T} = \{\mathbf{x}_1, \mathbf{x}_2, \ldots, \mathbf{x}_N\},
$$
where each data point $\mathbf{x}_t \in \mathbb{R}^d $ is a system state measured at discrete time $ t $. The sequence captures the temporal order of observations, reflecting how the state of the system evolves over time.
Particularly, when $d=1$, $\mathcal{T}$ degenerates into an univariate time series (UTS), when $d>1$, $\mathcal{T}$ refers $d$ dimensional multivariate time series (MTS).

\subsection{Problem Formulation}

Based on the preceding description, anomalies in multivariate time series from real-world systems often have no labels or very few. Consequently, our primary focus is on the challenge of unsupervised anomaly sequence detection.

\emph{Formulation 1 (Unsupervised Time Series Anomaly Detection)}
The \emph{time series model} typically exploits the temporal dependencies within the data to detect anomalies. Let $\mathcal{T} \in \mathbb{R}^{N \times d}$ represent a multivariate time series. The time series model is to learn a function $f(\mathcal{T})$, which takes the time series $\mathcal{T}$ as input and produces a binary output vector:

$$\mathbf{y} = f(\mathcal{T})$$

where $\mathbf{y} = [y_1, y_2, \dots, y_N]$, and each element $y_t \in \{0, 1\}$ indicates whether the state $\mathbf{x}_t$ at time $t$ is classified as normal ($y_t = 0$) or anomalous ($y_t = 1$) by a threshold $\eta$.

For any given test point $\mathbf{\hat{x}}_t$, where $t > N$. $N$ is the length of the training sequence. The TS deep learning model 
$\hat{f}(\mathcal{T})$ computes an anomaly score for the new data point $\mathbf{\hat{x}}_t$, where $f$ is a map of $\mathcal{T} \to y$. This score quantifies how much $\mathbf{\hat{x}}_t$ deviates from the distribution learned from the training data. By comparing this score to a threshold, the model determines whether the point should be classified as anomalous.

Our goal is to develop a TS model that incorporates a distinguishable criterion mechanism to generate anomaly scores which are both generalizable and discriminative,  so as to enable the identification of abnormal sequences without relying on explicit labels for training.

\section{Our Proposed Model}
\begin{figure*}[ht]
  \centering
  \includegraphics[width=\textwidth]{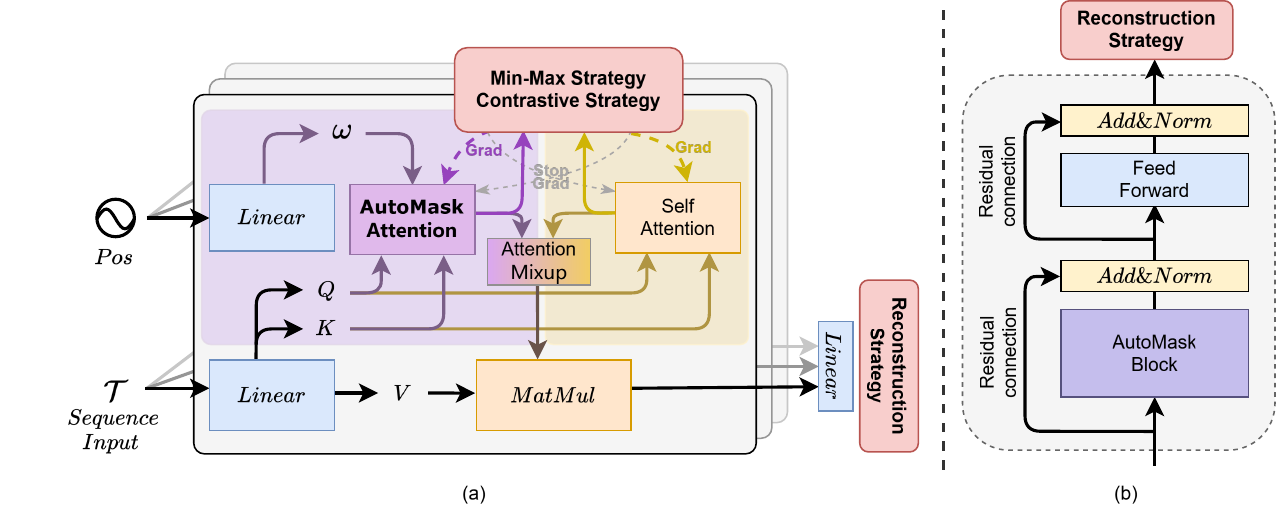}
  \caption{
  \textbf{Model Architecture} (a) \textbf{AutoMask Block}. %
  Our proposed AutoMask block integrates sinusoidal relative positional information while receiving $Q$ (queries) and $K$ (keys), forming a module similar to Rotary Position Embedding (RoPE) that encodes relative positional information across multiple learnable frequencies $ \omega $, thereby achieving the effect of learning multiscale sequence information (MSI). It also retains classical Self-Attention as a global sequence information (GSI)learner. After fusing the MSI and GSI through the Mixup module, attention weighting is applied to $V$ (values). The block is trained using a sequence reconstruction task. To ensure both Attention mechanisms effectively learn different aspects of sequence data information, we adopt a Max-Min training strategy under the global task of sequence reconstruction. (b) The position of the AutoMask block within the entire model stack, with an overall architecture consistent with that of the Transformer at the top level.}
  \label{fig:arch}
\end{figure*}

The theoretical foundation of our model primarily relies on empirical analogy and evaluation, involving studying how to design model architectures and well-crafted self-supervised tasks to train deep sequence representations with discriminative power. The effectiveness of the model is then validated through a series of evaluations on unsupervised sequence anomaly detection datasets, including comparative tests against baseline models.

In this section, we first introduce the overall architecture of our proposed AMAD model and provide a detailed explanation of the AutoMask Attention mechanism, which is designed to model multi-scale anomaly correlation structures, addressing the limitation of existing Transformer-based methods that fail to simultaneously account for
multi-scale
associations. We finally propose a Mixup-based attention fusion module to integrate local and global feature information. 

The next section outlines comprehensive training tasks used to train the model: the reconstruction task at the global level,the Max-Min strategy for local AutoMask feature extraction, and the contrastive strategy to align the co-sequence representations.

\subsection{Model Architecture} %

Given the classical Transformer and its variants in the field of sequence anomaly detection have not adequately addressed the aforementioned multi-scale learning capabilities, while maintaining the high-level structure consistent with the classical Transformer, we have redesigned the internal components of its basic blocks and propose our \textbf{A}uto\textbf{M}asked Attention for \textbf{A}nomaly \textbf{D}etection (AMAD) model, as illustrated in \autoref{fig:arch}.

The overall architecture of the model adheres to the structure of time series models. Unlike the classical Transformer, we retain only the encoder part to support the overarching sequence reconstruction task. Given an input $\mathcal{T} = \{\mathbf{x}_1, \ldots, \mathbf{x}_N\}$, the model consists of $L$ stacked identical AMAD modules. Each layer features a residual structure composed of AutoMask, LayerNorm, and FeedForward\cite{he2016deepresidual}, which has been proven effective in many previous works for adapting to sequence-related tasks and can mitigate gradient-related issues during training.

$$ \mathcal{H}_l = LayerNorm(\mathcal{T}_{l-1} + AutoMaskBlock(\mathcal{T}_{l-1})) $$
$$ \mathcal{T}_l = LayerNorm(\mathcal{S}_l + FeedForward(\mathcal{H}_l)) $$

Where $ \mathcal{H}_l, l \in \{1, \ldots, L\}$ denotes the intermediate hidden states. $\mathcal{T}_{l-1}$ and $\mathcal{T}_{l} \in \mathbb{R}^{N \times d}$ denote the model's $l$-th layer input and output separately.
We take GeLU \cite{hendrycks2016gelu} as our model's activation function. Particularly, the Feed Forward Layer can be defined as:

$$ FeedForward(x) = GeLU(Wx + b) $$

where $x$ represents the input vector, $W$ is the weight matrix, and $b$ is the bias vector.
The $ AutoMaskBlock $ computes and integrates information across different scales of the sequence data, which is called Cross-Attention Divergence.

The internal structure of the AutoMask block is illustrated on the left side of [fig:arch]. It incorporates positional information while processing $Q$ and $K$ simultaneously.
In practical models, the positional information $Pos$ is simplified as $\{1, 2, 3, \ldots, L\}$, where $L$ represents the length of the sequence. The parameter $\omega$ denotes multi-scale trigonometric frequencies learned based on positional information, modulating sequence information at different scales. AutoMask Attention fundamentally represents an improved version of Rotary Position Embedding (RoPE) that encodes relative positional information under multiple learnable frequencies $\omega$, enabling the learning of multiscale sequence information. 

The model retains the classical Self-Attention mechanism to act as a global information learner. Multiscale sequence information and global sequence information are fused through the Mixup module, after which attention weighting is applied to $V$ to generate the sequence output. The intermediate outputs from both Attention mechanisms are utilized for auxiliary task training.

The variables within the AutoMask block are outlined in \autoref{equ:attention1}, which includes four parts: block initialization, self-attention, AutoMask attention, and sequence reconstruction. Here, $\mathcal{X}^{l-1}$ denotes the output of the $l-1$th layer, while $W_{\mathcal{Q}}^{l}$, $W_{\mathcal{K}}^{l}$, $W_{\mathcal{V}}^{l}$, and $W_{\omega}$ represent the weight matrices for queries, keys, values, and autoencoder frequencies, respectively. $Pos$ is the position information vector, $d_{\text{model}}$ represents the model dimension, and $h$ denotes the number of heads in multi-head attention. $\mathcal{S}^{l}$ and $\mathcal{A}^{l}$ represent the outputs of self-attention and AutoMask attention, respectively. $\widehat{\mathcal{Z}}^{l}$ denotes the output of sequence reconstruction, and $\text{AttnMixup}$ indicates the output of attention fusion.

\begin{align}
    \text{Init:\ } \mathcal{Q},\mathcal{K},\mathcal{V} &= \mathcal{X}^{l-1}W_{\mathcal{Q}}^{l}, \mathcal{X}^{l-1}W_{\mathcal{K}}^{l}, \mathcal{X}^{l-1}W_{\mathcal{V}}^{l}, \\
    \mathbf{\omega} &= \mathcal{X}_{\omega} \cdot Pos \label{equ:attention1}\\
    \text{Self-Attn:\ } \mathcal{S}^{l} &= \mathrm{Softmax}\left(\frac{\mathcal{Q}\mathcal{K}^{\text{T}}}{\sqrt{d_{\text{model}}}}\right) \label{equ:attention2}\\
    \text{AutoMask-Attn:\ } \mathcal{A}^{l} &= \text{AutoMaskAttnBlock}(\mathcal{Q}^{l},\mathcal{K}^{l};\mathbf{\omega}) \label{equ:attention3}\\
    \text{Seq-Recon:\ } \widehat{\mathcal{Z}}^{l} &= \text{AttnMixup}(\mathcal{A}^{l},\mathcal{S}^{l})\mathcal{V} \label{equ:attention4}
\end{align}

In these equations, $\mathcal{Q}, \mathcal{K}, \mathcal{V} \in \mathbb{R}^{N \times d_{\text{model}}}$ and $\omega \in \mathbb{R}^{h \times 1}$, where $h$ is the number of attention heads. The multi-head attention mechanism in AMAD does not differ significantly from Transformer\cite{transformer} at the top-level architecture. In practice, position information $Pos$ is simplified to a static tensor $\{1, 2, 3, \ldots, N\}$.

The AMAD model characterizes anomaly correlation through \textbf{Cross-Attention Divergence (CAD)}. Specifically, this paper models the information gain between AutoMask Attention and Self-Attention using the Jensen-Shannon (JS) divergence, representing the correlation difference between multi-scale and full-scale representations, termed as Cross-Attention Divergence (CAD).

Due to the asymmetry of KL divergence, we cannot directly use KL divergence or its average; otherwise, there might be potential misalignment issues. Therefore, we choose the symmetric JS divergence. JS divergence first defines the average distribution of two distributions as an anchor point, then computes the average KL divergence of the two distributions relative to this average distribution. This anchor point acts as a boundary line, inspiring the Max-Min strategy used for training the model, detailed in the following sections. The definition of Cross Attention  Divergence (CAD) is given in \autoref{eq:cad}:

\begin{equation}
    \mathrm{CAD}(\mathcal{A},\mathcal{S};\mathcal{X}) = \left[\frac{1}{L}\sum_{l=1}^{L}\mathrm{D}_{JS}(\mathcal{A}_{i,:}^{l}\|\mathcal{S}_{i,:}^{l})\right]_{i=1,\cdots,N}
    \label{eq:cad}
\end{equation}

Here, $\mathrm{CrossAttention Divergence}$ (CAD) represents the correlation difference between AutoMask Attention and Self-Attention, with $\mathcal{A}$ and $\mathcal{S}$ defined as the outputs of the two attention modules. $\mathrm{CAD}(\mathcal{A},\mathcal{S};\mathcal{X}) \in \mathbb{R}^{N \times 1}$, where $N$ is the sequence length, $L$ is the number of layers in the model, $\mathcal{A}_{i,:}^{l}$ denotes the $i$th time-series data point of the $l$th layer's AutoMask Attention, and $\mathcal{S}_{i,:}^{l}$ denotes the $i$th time-series data point of the $l$th layer's Self-Attention. The computation of CAD involves calculating the JS divergence between AutoMask Attention and Self-Attention at each position and then averaging these values.

The AutoMask attention mechanism (\texttt{AutoMaskAttnBlock}) and the attention fusion module (\texttt{Mixup}) will be elaborated in subsequent sections.

\subsection{AutoMask Attention}

Before introducing our AutoMask attention mechanism, we summarize that existing deep unsupervised sequence anomaly detection models can be categorized as attempts to model certain types of relative relationships $g(\cdot)$ within sequences, as shown in \autoref{tab:anoasso}.

\begin{table}[!htb]
  \centering
  \caption{Relational Modeling in Sequential AD}
  \begin{tabularx}{.48\textwidth}{p{3cm} l}
  \toprule
  \textbf{Modeling Approach} & \textbf{Relational Model} \\ \midrule
  Reconstruction (single point) & $f(x_t) = g(\hat{x}_t; t)$ \\
  Interpolation (random association) & $f(x_{t-1}, x_{t+1}) = g(\hat{x}_t; RandomNoise)$ \\
  Co-relation (sequence association) & $f(x_{t-n}, \ldots, x_{t+n}) = \langle f_{q}(\mathbf{\alpha}_{m},m),f_{k}(\mathbf{\alpha}_{n}, n)\rangle$ \\
  Kernel (e.g. gaussian) & $f(x_m, x_n) =
    \frac{1}{\sqrt{2\pi}\sigma_m}\exp\left(-\frac{|n-m|^2}{2\sigma_m^2}\right)$ \\ 
  \bottomrule
  \end{tabularx}

  \label{tab:anoasso}
\end{table}

Existing deep unsupervised sequence anomaly detection models can be summarized as attempts to model certain types of relative relationships within sequences.

\textbf{Sequence Reconstruction} utilizes sequence reconstruction for single-point modeling, aiming to reconstruct the input sequence to detect anomalies. Such as USAD\cite{usadkdd20}, TranAD\cite{tuli2022tranad}, etc.

\textbf{Random Interpolation} employs random mask and interpolation, focusing on modeling the associated anomalies in sequences under noise perturbations, such as ImDiffusion \cite{chen2023imdiffusion}.

\textbf{Sequence-wide Correlation} aims to establish sequence-wide correlations. Such as Transformer\cite{transformer,su2019ominianomaly} and many varients, which captures long-range dependencies between sequence elements by doing the matrix inner products.

\textbf{Kernel Method} enhances sensitivity to local patterns by incorporating a learnable kernel function. Such as gaussian kernael in AnomalyTransformer\cite{xu2021anomalytrans}. However gaussian kernel is not enough to capture the multi-scale dependencies.

The Fourier decomposition process provides the key insight that decomposing any function using orthogonal sine and cosine functions at multiple frequencies. The relationship between exponentials and trigonometric functions, established by Euler's formula (\autoref{eq:euler}), leads us to the Fourier series representation of an arbitrary function $ f $ as given in \autoref{eq:fourier}. In these equations, $ c_n $ represents the Fourier coefficients, $ \omega_n $ denotes the frequency, and $ i $ is the imaginary unit. The core idea of Fourier Decomposition is to express any function as a sum of trigonometric functions at different frequencies, which can be converted into exponential form based on Euler's formula.

This insight inspires us to explore a generalized correlation model distinct from previous methods, taking the form shown in \autoref{eq:aminsight}:

\begin{align}
    \text{Euler's Formula:\ } & e^{i\theta} = \cos \theta + i\sin \theta
    \label{eq:euler} \\
    \text{Fourier Series:\ } & f(x) = \sum_{n=-\infty}^{\infty} c_n e^{i\omega_n x}
    \label{eq:fourier} \\
    \text{AutoMask (Prototype):\ } & f(x) = \underset{\omega}{\text{Operation}\ } g(\cdot; \omega)
    \label{eq:aminsight} 
\end{align}

\begin{figure*}[!htb]
  \centering
  \includegraphics[width=.85\textwidth]{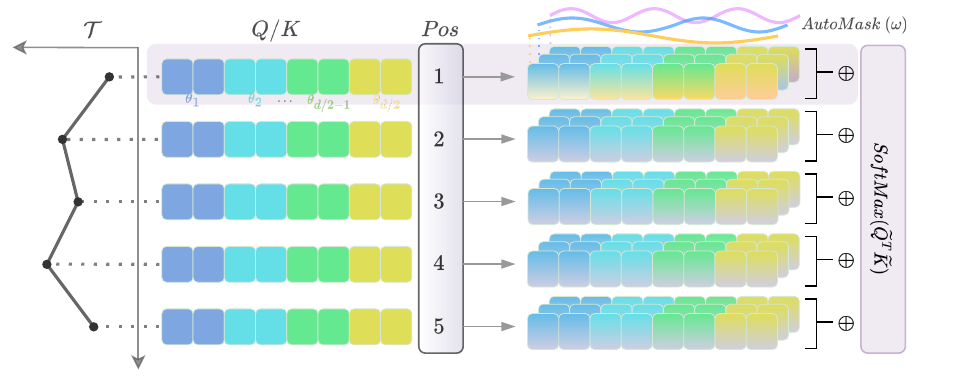}
  \caption{
  The AutoMask attention Mechanism incorporates a learnable modulation mechanism, which we named \emph{AutoMask}. This mechanism directly draws inspiration from the idea of Fourier Decomposition, where a sufficient number of orthogonal trigonometric functions can combine to form any curve. In contrast to Rotary Position Embedding (RoPE)\cite{su2024roformer}, AutoMask introduces multiple learnable trigonometric modulation terms with dynamic weights denoted by $ \omega $. We refer to this as \emph{Automatic Masking}.
  On the basis of learnable automatic masking, we employ Max-Min and contrastive training strategies (explained in the context) to make AutoMask more inclined to fit local features of the sequence. Each pair of components in the $ Q/K $ vectors corresponds to a RoPE rotation angle $ \theta $, embedding absolute positional information. After $ l $ embeddings, the resulting vectors undergo weighted linear combination through AutoMask, yielding the AutoMask-embedded vectors $ \widetilde{Q} $ and $ \widetilde{K} $.
  AutoMask Attention}
  \label{fig:amattn}
\end{figure*}

\emph{AutoMask attention} is our implementation of the correlation model described in \autoref{eq:aminsight}. Essentially, it dynamically modulates sequence information at different scales under learnable rotation masks constructed from multiple learnable frequencies $ \omega $, embedding relative positional information. This process is an improvement over RoPE (Rotary Position Embedding)\cite{su2024roformer}, designed to automatically adjust multiple rotation masks to represent sequences and focus on multi-scale information. The workflow is illustrated in \autoref{fig:amattn}, where $ \oplus $ denotes the multi-head concatenation operation.

The AutoMask attention mechanism accepts inputs $\mathcal{Q}$, $\mathcal{K}$, and frequency vectors $\omega$. We first define the \emph{rotary mask embedding} for multi-head $\mathcal{Q}$ and $\mathcal{K}$ tensors as shown in \autoref{eq:amarope}, which is a learnable variation of RoPE\cite{su2024roformer}:

\begin{equation}
    \begin{split}
    \widetilde{\mathcal{Q}}^\eta &= f^\eta_\mathcal{Q}(\mathcal{Q}_n, n) = (W_\mathcal{Q}^\eta \mathcal{Q}_n^\eta)e^{in\omega_{\eta}\theta}, \quad \eta \in \{1, \ldots, h\} \\
    \widetilde{\mathcal{K}}^\eta &= f^\eta_\mathcal{K}(\mathcal{K}_m, m) = (W_\mathcal{K}^\eta \mathcal{K}_m^\eta)e^{im\omega_{\eta}\theta}, \quad \eta \in \{1, \ldots, h\}
    \end{split}
    \label{eq:amarope}
\end{equation}

In these equations, $\theta \in \mathbb{R}$ is a fixed rotation angle parameter, $n$ and $m$ represent the position indices of the sequence, and $h$ denotes the number of attention heads. For clarity, the function $f$ can be written in the form of a rotation matrix modulated by $\omega$, as shown in \autoref{eq:amaropemat}, which is equivalent to the above equations.

\begin{equation}
    f^\eta(x, p) = 
    \begin{pmatrix} 
        \cos(p\omega_\eta\mathbf{\theta}) & -\sin(p\omega_\eta\mathbf{\theta}) \\ 
        \sin(p\omega_\eta\mathbf{\theta}) & \cos(p\omega_\eta\mathbf{\theta})
    \end{pmatrix}
    \begin{pmatrix}
        x_{2\kappa} \\ x_{2\kappa+1}
    \end{pmatrix}, \quad \kappa \in \{0, \ldots, d/2\}
    \label{eq:amaropemat}
\end{equation}

Here, $x_{2\kappa}$ and $x_{2\kappa+1}$ represent the even and odd indexed elements of the input tensor $x$ along the feature dimension, $p$ represents the position index of the input tensor, $\omega_r$ denotes the $r$-th frequency, and $\mathbf{\theta}$ represents the rotation angle.

The fundamental difference between the rotary mask embedding in AutoMask and the Rotary Position Embedding (RoPE) \cite{su2024roformer} lies in how they embed relative positional information. RoPE uses fixed rotation angles $\theta$ and position information $p$, whereas AutoMask employs learnable frequencies $\omega$ and position information $p$. This allows AutoMask to dynamically modulate sequence information at different scales, embedding relative positional representations.

Next, through concatenation (\texttt{Concat}) operations, we obtain the complete query $\widetilde{\mathcal{Q}}$ and key $\widetilde{\mathcal{K}}$ tensors after rotary mask embedding, as shown in \autoref{eq:amaropeconcat}:

\begin{equation}
    \begin{split}
    \widetilde{\mathcal{Q}} &= \text{Concat}(\widetilde{\mathcal{Q}}^1, \ldots, \widetilde{\mathcal{Q}}^h) = f_\mathcal{Q}(\mathcal{Q}_m, m) = (W_\mathcal{Q} \mathcal{Q}_n)e^{in\Omega\theta}\\
    \widetilde{\mathcal{K}} &= \text{Concat}(\widetilde{\mathcal{K}}^1, \ldots, \widetilde{\mathcal{K}}^h) = f_\mathcal{K}(\mathcal{K}_m, m) = (W_\mathcal{K} \mathcal{K}_n)e^{in\Omega\theta} 
    \end{split}
    \label{eq:amaropeconcat}
\end{equation}

Here, $\Omega$ represents the concatenated multi-head rotation frequencies ($\omega$s).

\label{sec:amaasso}

Based on this, we can define a relative positional correlation function $g(\cdot)$, which embeds multi-scale relative positional information. We name this correlation \emph{AutoMask} correlation, as shown in \autoref{eq:ama_g}:

\begin{equation}
    g(\mathcal{Q}, \mathcal{K}; \omega) = \widetilde{\mathcal{Q}} \widetilde{\mathcal{K}}^{\text{T}}
    \label{eq:ama_g}
\end{equation}

Thus, we derive the specific expression for the AutoMask correlation model as shown in \autoref{eq:aminsight2}, which conforms to our expected prototype, as given in \autoref{eq:aminsight}. By comparing the specific AutoMask correlation with the Fourier series in \autoref{eq:fourier}, we observe that the use of multiple frequency-based rotary mask embeddings structurally mirrors the idea of decomposing arbitrary functions using sine and cosine functions. This leads us to a generalized correlation model distinct from previous methods. Through AutoMask correlation, our model can learn true multi-scale information representations of sequences, addressing the multi-scale correlation representation problem mentioned earlier. 

\begin{equation}
\text{AutoMask (Model):\ } f(x) = \underset{\omega}{\text{Concat}\ } g(\mathcal{Q}, \mathcal{K}; \omega)
\label{eq:aminsight2} 
\end{equation}

Here, the function $g$ is defined as in \autoref{eq:ama_g}, and $\omega$ represents the learnable frequency parameters.  It is worthy to notify that the $\omega$ here does not need to satisfy orthogonality, as our objective is to learn the multiscale information of the sequence rather than enforce complete orthogonality.

We can then define the AutoMask attention mechanism, where the specific mathematical expression of \autoref{equ:attention3} is given in \autoref{eq:amattn}:

\begin{equation}
  \begin{split}
    \mathcal{A} &= \text{AutoMaskAttnBlock}(\mathcal{Q}, \mathcal{K}; \omega) \\
    &= \text{Softmax}\left(\frac{g(\mathcal{Q}, \mathcal{K}; \omega)}{\sqrt{d_{q} \cdot d_{k}}}\right) \\
    &= \text{Softmax}\left(\frac{\widetilde{\mathcal{Q}} \widetilde{\mathcal{K}}^{\text{T}}}{\sqrt{d_{q} \cdot d_{k}}}\right)
  \end{split}
  \label{eq:amattn}
\end{equation}

Here, $d_{q}$ and $d_{k}$ denote the dimensions of the query ($\mathcal{Q}$) and key ($\mathcal{K}$), respectively, which are equal to the model dimension $d_{\text{model}}$ in this context. The tensors $\widetilde{\mathcal{Q}}$ and $\widetilde{\mathcal{K}}$ are the query and key tensors with rotary mask embeddings, as described earlier.

\subsection{Attention Fusion}

The attention fusion module is implemented through a Mixup operation, which essentially performs an attention-weighted fusion with a hyperparameter. Its mathematical form is given in \autoref{eq:attnmixup}:

\begin{equation}
    \text{AttnMixup}(\mathcal{A}, \mathcal{S}) = \alpha \mathcal{A} + (1-\alpha) \mathcal{S}
    \label{eq:attnmixup}
\end{equation}

Here, $\alpha$ is a hyperparameter that controls the degree of fusion between the two attention modules. In the parameter search section of our experiments, we will discuss in detail how the selection of this hyperparameter affects model performance.

\section{Training Strategies}

The key to the sequence representation problem lies in designing reasonable self-supervised proxy strategies. We have designed two self-supervised strategies—the Max-Min strategy and the multi-scale contrastive representation alignment strategy—combined with a global proxy task of sequence reconstruction to train the deep sequence model AMAD proposed in this chapter.

The global loss can be defined as shown in \autoref{eq:loss}, which is divided into two stages: the reconstruction stage and the contrastive stage, corresponding to the two self-supervised strategies.

\begin{equation}
    \begin{split}
    \text{Reconstruction Stage:\ }&\mathcal{L}_{\text{Recon}}({\widehat{\mathcal{X}}},\mathcal{A},\mathcal{S},\lambda;\mathcal{X})=\|\mathcal{X}-{\widehat{\mathcal{X}}}\|_{\text{F}}^2\\
    & -\lambda\times \|\mathrm{CAD}(\mathcal{A},\mathcal{S};\mathcal{X})\|_{1} \\
    \text{Contrastive Stage:\ }&\mathcal{L}_{\text{Contrastive}} \\
    \text{Training Loss:\ }&\mathcal{L}{\text{Total}} = \mathcal{L}_{\text{Recon}} + \mathcal{L}_{\text{Contrastive}}
    \end{split}
    \label{eq:loss}
\end{equation}

Here, $\|\cdot\|_{\text{F}}$ denotes the Frobenius norm, $\lambda$ is a hyperparameter used to adjust the loss weight, $\mathcal{L}_{\text{Contrastive}}$ represents the contrastive representation loss, which will be defined later, and $\mathrm{CAD}(\mathcal{A},\mathcal{S};\mathcal{X})$ is the Cross-Attention Divergence defined in \autoref{eq:cad}.

Both strategies attempt to establish multi-scale and local correlation information for sequences through the AutoMask attention mechanism and global correlation information through Self-Attention by designing representation losses, thereby achieving multi-scale representation of sequences. The Max-Min strategy uses the Cross-Attention Divergence (CAD) defined in \autoref{eq:cad} as a correlation metric to optimize the direction and construct correlation difference loss; the multi-scale contrastive representation alignment strategy constructs contrastive representation loss by designing contrastive sample pairs.

\subsection{Max-Min Strategy}

\begin{figure}[!htb]
  \centering
  \includegraphics[width=.5\textwidth]{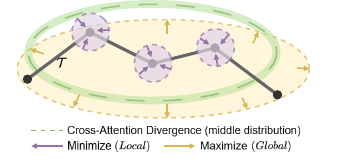}
  \caption{
  The Max-Min strategy. Steering AutoMask Attention to primarily represent local features, while Self Attention focuses on the global characteristics of the sequence. This is achieved by constructing a prior mean distribution based on the Cross-Attention Divergence defined using JS divergence as an anchor point (the green circular area). Both Attention outputs are intermediate logits. The minimization step updates only the weights of the AutoMask Attention sub-module, and the maximization step updates only the weights of the Self Attention sub-module.
  By reducing the correlation between the AutoMask Attention logits and the intermediate distribution and increasing the correlation between the Self Attention logits and the intermediate distribution, we can generally conclude that AutoMask Attention will focus more on local features of the sequence (the purple part), while Self Attention, as expected, will focus more on the overall features of the sequence (the light yellow part).
  Max-Min Strategy}
  \label{fig:Max-Min}
\end{figure}

We first provide the definition of the training loss for the Max-Min strategy, as shown in \autoref{eq:Max-Minloss}:

\begin{equation}\label{eq:Max-Minloss}
    \begin{split}
   \text{Minimization Phase (Min):}&\ \mathcal{L}_{\text{Recon}}({\widehat{\mathcal{X}}},\mathcal{A},\mathcal{S}_{\text{detach}},-\lambda;\mathcal{X})\\ 
   \text{Maximization Phase (Max):}&\ \mathcal{L}_{\text{Recon}}({\widehat{\mathcal{X}}},\mathcal{A}_{\text{detach}},\mathcal{S},\lambda;\mathcal{X})
    \end{split}
\end{equation}

Here, $\lambda > 0$ and $\text{detach}$ indicates stopping the backward propagation of gradients.

In the Max-Min strategy represented by \autoref{eq:Max-Minloss}, we aim to have AutoMask Attention primarily converge on local feature representation during the minimization phase, while Self Attention focuses more on the overall sequence features during the maximization phase. The intuitive principle behind this is that the Cross-Attention Divergence defined in \autoref{eq:cad} is based on JS divergence. JS divergence relies on constructing a mean distribution as an anchor point (the green box in \autoref{fig:Max-Min}), with the two attention outputs $\mathcal{A}$ and $\mathcal{S}$ serving as logits. This aligns with the computational path connected by the strategy blocks in \autoref{fig:arch}. The Max-Min strategy updates only the weights of the AutoMask Attention sub-module during the minimization phase and only the weights of the Self Attention sub-module during the maximization phase.

Intuitively, we can summarize the optimization direction achieved by the Max-Min strategy as illustrated in \autoref{fig:Max-Min}. Specifically, by reducing the correlation between the AutoMask Attention logits $\mathcal{A}$ and the intermediate distribution and increasing the correlation between the Self Attention logits $\mathcal{S}$ and the intermediate distribution, this strategy ensures that AutoMask Attention focuses more on local features of the sequence (the purple part in \autoref{fig:Max-Min}), while Self Attention, as expected, focuses more on the overall features of the sequence (the orange part in \autoref{fig:Max-Min}).

It is worth noting that similar strategies are adopted by many methods but with different training objectives. Such strategy has been proved effective in preventing the model from descending to a trivial solution in AnomalyTransformer\cite{xu2021anomalytrans},USAD\cite{usadkdd20}, TrainAD\cite{tuli2022tranad}. In detail, Anomaly Transformer utilizes Minimax strategy in  to mitigate the degeneration of Gaussian kernels. USAD and TrainAD designed an similar adversarial loss to mimic small perturbations, thus making the model more sensitive to the less significant anomaly. 
Whereas, our Max-Min strategy fundamentally differs from the Minimax strategy used in others, which aims to empower the model to learn differential representations of local and global sequence characteristics.

\subsection{Local-Global Contrastive Strategy}

\begin{figure}[!htb]
  \centering
  \begin{lstlisting}[language=Python]
  # input_data[B, L, D] Input data
  # self_series[B, H, W, E] Logits generated by the Self Attention mechanism
  # ama_series[B, H, W, E] Logits generated by the AutoMask Attention mechanism
  # t Temperature coefficient
  
  # Initialize contrastive alignment loss
  con_align_loss = 0.0
  # Create alignment labels representing each sample's index
  align_labels = torch.arange(B).to(self.device)
  # Iterate over sequences generated by the Self Attention mechanism
  for u in range(len(ama_series)):
      # Flatten global and local features into two-dimensional matrices
      l_global = self_series[u].view(B, -1)
      l_local = ama_series[u].view(B, -1)
      # Compute alignment logits using matrix multiplication and exponential of the temperature coefficient
      logits_align = torch.mm(l_global, l_local.t()) * np.exp(t)
      # Calculate cross-entropy loss and add it to the contrastive alignment loss
      con_align_loss += F.cross_entropy(logits_align, align_labels)
  # Compute the final contrastive loss by averaging the contrastive alignment loss
  contrastive_loss = con_align_loss / B
  
  \end{lstlisting}
  \caption{Key Code for Contrastive Alignment Strategy}
  \label{fig:chap4:contrastive_loss_code}
\end{figure}

\begin{figure}[!htb]
  \centering
  \includegraphics[width=0.48\textwidth]{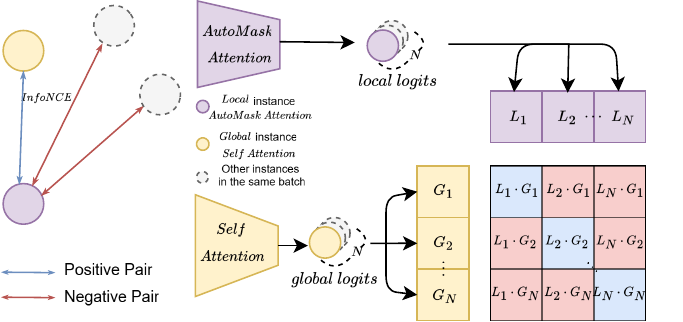}
  \caption{
  The Local-Global Contrastive Strategy.
  On the left side is an illustration of positive and negative example pairs for the instance discrimination task (Individual Discrimination). On the right side, the specific workflow of the instance discrimination task is depicted. This task jointly considers the AutoMask attention logits and self attention logits corresponding to \(N\) sequences within the same batch as dual sample pairs, where sequences from the same origin form positive pairs, and others form negative pairs, used to align the local and global representations of the same sequence.
  Building on the Max-Min mechanism that selectively learns multi-scale features of local and global aspects, we intuitively notice the alignment issue between local and global feature representations. To address this problem, we have designed a Local-Global contrastive training strategy to align local and global representations. Specifically, the Local-Global contrastive strategy employs an instance discrimination proxy task, treating the AutoMask logits and self attention logits corresponding to \(N\) sequences in a batch as dual sample pairs and computing their pairwise dot products. Sample pairs from the same sequence are considered positive examples, while all other pairs are negative examples. We use InfoNCE as the contrastive loss. The contrastive strategy adds constraints on internal consistency within sequence samples, based on the focus of the Max-Min strategy on local and global aspects of sequences, thereby aligning the local and global representations of the same sequence.
}
  \label{fig:chap4:contrastive}
\end{figure}

The Max-Min mechanism proposed in the previous section can learn multi-scale representations that distinguish between local and global features, but it ignores the multi-scale relationships within sequences, i.e., it does not address the alignment of feature representations within the same sequence. To solve this problem, we design a Local-Global contrastive training strategy in this section to align local and global representations. Specifically, the Local-Global contrastive strategy employs an instance discrimination proxy task, treating the AutoMask logits and Self Attention logits corresponding to $N$ sequences in a batch as dual sample pairs and computing their pairwise dot products. Sample pairs from the same sequence are considered positive examples, while all the other pairs are negative examples. We use a varient of InfoNCE loss\cite{oord2018infonce} as the contrastive loss. This contrastive strategy adds constraints on internal consistency within sequence samples, based on the focus of the Max-Min strategy on local and global aspects of sequences, thereby aligning the local and global representations of the same sequence.

The multi-scale contrastive representation alignment strategy proposed in this chapter is based on the instance discrimination task in contrastive learning \cite{wu2018instdist}, but borrows the architecture form from CLIP \cite{clip}, as shown in \autoref{fig:chap4:contrastive}. On the left side, there is an illustration of positive and negative example pairs for the instance discrimination task (Individual Discrimination). On the right side, the specific workflow of the instance discrimination task is depicted, where the task utilizes AutoMask Attention logits $\mathcal{A}$ and Self Attention logits $\mathcal{S}$ corresponding to $B$ time-series data in the same batch as dual sample pairs, with sequences forming positive pairs and others forming negative pairs. This strategy is used to align the local and global representations of the same sequence.

Based on the aforementioned strategy, we define the training logits $l$ and instance labels $y$ for the multi-scale contrastive representation alignment strategy as follows (\autoref{eq:chap4:contrastive_logits}):

\begin{equation}
    \begin{split}
    l &= \dot{\mathcal{S}}\dot{\mathcal{A}}^\text{T} \exp(\tau) \\
    y &= {0, 1, \ldots, B-1}
    \end{split}
    \label{eq:chap4:contrastive_logits}
\end{equation}

Here, $\dot{\mathcal{S}}$ and $\dot{\mathcal{A}}$ represent the Attention logits $\mathcal{S}$ and $\mathcal{A}$ reshaped to $[B, -1]$.

We implement a variant of the InfoNCE contrastive loss specific to our contrastive representation task using cross-entropy loss, equivalent to the InfoICE Loss\cite{oord2018infonce}.
\begin{equation}
    \mathcal{L}_{\text{Contrastive}} = \text{CrossEntropy}(l, y)
    \label{eq:chap4:contrastive_loss}
\end{equation}
where $\tau$ is a temperature hyperparameter used to adjust the scale of the contrastive loss. The key code for implementing the contrastive alignment strategy is shown in \autoref{fig:chap4:contrastive_loss_code}.

\section{Experiments}

This section first introduces the datasets used in this paper, which are widely employed to evaluate the performance of sequence anomaly detection algorithms.

\subsection{Anomaly Discrimination Method}

\label{sec:amadmetric}

Based on the model we formally described above, we define the anomaly discrimination score (Anomaly Score) as:
\begin{equation}
\begin{split}
\mathrm{AnomalyScore}(\mathcal{X}) =~& \mathrm{Softmax}\Big(-\mathrm{CAD}(\mathcal{A},\mathcal{S};\mathcal{X})\Big)\\ &\odot\Big[\|\mathcal{X}_{i,:}-{\widehat{\mathcal{X}}}_{i,:}\|_{\text{2}}^2\Big]_{i=1,\cdots,N}
\end{split}
\label{equ:metric}
\end{equation}

Here, $\mathcal{X}$ is the input sequence, $\mathcal{A}$ and $\mathcal{S}$ are the outputs of AutoMask and Self Attention respectively, and $\mathrm{CAD}$ is the co-attention divergence, all of which have been explained earlier. $\widehat{\mathcal{X}}$ is the reconstructed sequence, $\odot$ denotes the Hadamard product, and $\|\cdot\|_{\text{2}}$ represents the L2 norm.

Our anomaly discrimination score is based on the product of reconstruction error and the opposite of co-attention divergence. This approach leverages the multi-scale correlation difference information modeled by the product to weight the reconstruction, making it suitable as a criterion for anomaly detection. Thresholding the $p$-th percentile of the $\mathrm{AnomalyScore}$ enables sequence abnormality detection. The percentile $p$ is a dataset - preset parameter provided as the ar value later, and it does not equal the actual anomaly ratio of the dataset.

Noting that former works adopted multitude of anomaly score threshold pick methods. For example, POT and SPOT\cite{potspot} were classical methods adopted by USAD and TranAD, but which need calibration steps and should pick initial parameters carefully. Anomaly-Transformer\cite{xu2021anomalytrans} and ImDiffusion\cite{chen2023imdiffusion} utilized the percentile of the anomaly measurement, which needs a prior estimation of the anomaly. We follow the latter method, and the prior anomaly ratio is illustrated in \autoref{tab:amad_results}, which is the same as AnomalyTransformer.
The percentile only method utilized by recent methods fixed the $ar$ hyperparameter for fair comparison, as the anomaly score is directly calculated from the model output.

Our prebidicted anomaly type is 
$$ Y = \mathbbm{1}_{\mathrm{AnomalyScore}(\mathcal{X}) > \mathrm{Percentile}(\mathrm{AnomalyScore}(\mathcal{X}),p)},$$
where $\mathbbm{1}$ is the indicator function (equal to 1 if the condition is met, indicating an anomaly), and $\mathrm{Percentile}$ is the percentile function.

The model we proposed are trained and tested in a PyTorch environment with the following specifications: an NVIDIA RTX 4090D $\times 1$ GPU, 128GB RAM, 2TB SSD, Intel Xeon E5-2699 v5 CPU, Ubuntu 20.04, PyTorch 1.9.0, CUDA 11.1, cuDNN 8.0.5, and Python 3.8.5. The learning rate is set to $lr=0.02$ with an exponential decay strategy, a batch size of 256, and early stopping (patience=3) to prevent overfitting.

Other hyperparameters for the model are listed in \autoref{tab:amadparam}.

\begin{table}[htbp]
    \centering
    \caption{AMAD Model Hyperparameters}
      \begin{tabularx}{0.5\textwidth}{>{\centering \arraybackslash}X|>{\centering \arraybackslash}X}
      \toprule
      Hyperparameter & Value \\
      \midrule
      layers & 3 \\
      model dimension & 512 \\
      attention heads & 8 \\
      $\lambda$ (Loss parameter) & 3 \\
      epochs & $\le$ 10 \\
      \bottomrule
      \end{tabularx}%
    \label{tab:amadparam}%
\end{table}%

\begin{table*}[!htb]
  \centering
  \caption{Experimental Results for AMAD and Baseline Methods 
}
  \scalebox{0.82}{
\begin{tabular}{c|ccc|ccc|ccc|ccc|ccc}
  \toprule
  \multicolumn{1}{c|}{Dataset} & \multicolumn{3}{c|}{MSL ($ar=1$)} & \multicolumn{3}{c|}{SWaT ($ar=1$)} & \multicolumn{3}{c|}{PSM ($ar=1$)} & \multicolumn{3}{c|}{SMAP ($ar=1$)} & \multicolumn{3}{c}{SMD ($ar=0.5$)} \\
  \midrule
  \multicolumn{1}{c|}{Metric} & \multicolumn{1}{c}{P} & \multicolumn{1}{c}{R} & \multicolumn{1}{c|}{F1} & P   & \multicolumn{1}{c}{R} & \multicolumn{1}{c|}{F1} & \multicolumn{1}{c}{P} & \multicolumn{1}{c}{R} & \multicolumn{1}{c|}{F1} & \multicolumn{1}{c}{P} & \multicolumn{1}{c}{R} & \multicolumn{1}{c|}{F1} & P   & \multicolumn{1}{c}{R} & \multicolumn{1}{c}{F1} \\
  \midrule
  MAD-GAN\cite{li2019madganma} & 0.8157 & 0.9216 & 0.8654 & \multicolumn{1}{c}{0.7918} & 0.5423 & 0.6385 & 0.8596 & 0.8838 & 0.8698 & \underline{0.9547} & 0.5474 & 0.6952 & \multicolumn{1}{c}{0.8851} & 0.9045 & 0.8803 \\
  THOC\cite{THOC} & 0.8845 & 0.9097 & 0.8969 & \multicolumn{1}{c}{0.8394} & 0.8636 & 0.8513 & 0.8814 & 0.9099 & 0.8954 & 0.9206 & 0.8934 & 0.9068 & \multicolumn{1}{c}{0.7976} & 0.9095 & 0.8499 \\
  InterFusion & 0.8128 & 0.9270 & 0.8662 & \multicolumn{1}{c}{0.8059} & 0.8558 & 0.8301 & 0.8361 & 0.8345 & 0.8352 & 0.8977 & 0.8852 & 0.8914 & \multicolumn{1}{c}{0.8702} & 0.8543 & 0.8622 \\
  BeatGAN\cite{zhou2019beatganar} & 0.8975 & 0.8542 & 0.8753 & \multicolumn{1}{c}{0.6401} & \multicolumn{1}{c}{\underline{0.8746}} & 0.7392 & 0.9030 & 0.9384 & 0.9204 & 0.9238 & 0.5585 & 0.6961 & \multicolumn{1}{c}{0.7290} & 0.8409 & 0.7810 \\
  DAGMM\cite{zong2018dagmm} & 0.8960 & 0.6393 & 0.7462 & \multicolumn{1}{c}{0.8992} & 0.5784 & 0.7040 & 0.9349 & 0.7003 & 0.8008 & 0.8645 & 0.5673 & 0.6851 & \multicolumn{1}{c}{0.6730} & 0.4989 & 0.5730 \\
  MTAD-GAT\cite{zhao2020mtadgat} & 0.7321 & 0.7616 & 0.7432 & \multicolumn{1}{c}{0.8468} & 0.8224 & 0.8344 & 0.8763 & 0.8725 & 0.8744 & \textbf{0.9718} & 0.5259 & 0.6824 & \multicolumn{1}{c}{0.8836} & 0.8330 & 0.8463 \\
  LSTM-AD\cite{malhotra2015lstmad} & 0.7330 & 0.5745 & 0.6378 & \multicolumn{1}{c}{\textbf{0.9925}} & 0.6737 & 0.8026 & 0.9050 & 0.7707 & 0.8313 & 0.7841 & 0.5630 & 0.6554 & \multicolumn{1}{c}{0.3361} & 0.3229 & 0.2639 \\
  OmniAnomaly\cite{su2019ominianomaly} & 0.8902 & 0.8637 & 0.8767 & \multicolumn{1}{c}{0.8142} & 0.8430 & 0.8283 & 0.8839 & 0.7446 & 0.8083 & 0.9249 & 0.8199 & 0.8692 & \multicolumn{1}{c}{0.8368} & 0.8682 & 0.8522 \\
  TranAD\cite{tuli2022tranad} & 0.8951 & 0.9297 & 0.9115 & \multicolumn{1}{c}{0.7025} & 0.7266 & 0.6886 & 0.9506 & 0.8951 & 0.9220 & 0.8224 & 0.8502 & 0.8361 & \multicolumn{1}{c}{0.8906} & 0.8982 & 0.8785 \\
  AnomalyTransformer\cite{xu2021anomalytrans} & \textbf{0.9209} & \multicolumn{1}{c}{\underline{0.9515}} & \multicolumn{1}{c|}{\underline{0.9359}} & \multicolumn{1}{c}{0.9155} & \textbf{0.9673} & \textbf{0.9407} & 0.9691 & \multicolumn{1}{c}{\underline{0.989}} & \multicolumn{1}{c|}{\underline{0.9789}} & 0.9413 & \multicolumn{1}{c}{\underline{0.9870}} & \multicolumn{1}{c|}{\underline{0.9636}} & \multicolumn{1}{c}{0.8940} & \textbf{0.9545} & \multicolumn{1}{c}{\underline{0.9233}} \\
  ImDiffusion\cite{chen2023imdiffusion} & 0.8930 & 0.8638 & 0.8779 & \multicolumn{1}{c}{0.8988} & 0.8465 & \multicolumn{1}{c|}{\underline{0.8709}} & \textbf{0.9811} & 0.9753 & 0.9781 & 0.8771 & 0.9618 & 0.9175 & \multicolumn{1}{c}{\textbf{0.9520}} & \multicolumn{1}{c}{\underline{0.9509}} & \textbf{0.9488} \\
  \midrule
  AMAD (ours) & \underline{0.9190} & \textbf{0.9569} & \textbf{0.9375} & \underline{0.9844} & 0.7134 & 0.8273 & \underline{0.9726} & \textbf{0.9910} & \textbf{0.9817} & 0.9432 & \textbf{0.9948} & \textbf{0.9683} & \underline{0.9043} & 0.8471 & 0.8748 \\
  \bottomrule
  \end{tabular}%
}%
  \label{tab:amad_results}%
\end{table*}%

\subsection{Benchmark Datasets}

The evaluation of our method's effectiveness uses five publicly available sequence anomaly detection datasets: MSL, SWaT, PSM, SMAP, and SMD. These datasets consist of real-world data collected from various domains including network traffic, server operations, aerospace networks, and water treatment networks, and they are widely used to assess the performance of self-supervised and unsupervised anomaly detection algorithms.

These datasets contain a substantial amount of normal data and a smaller proportion of anomalous data. Note that only the test set data is labeled for evaluating algorithm performance.

Below is a brief introduction to each of these datasets:

\textbf{Server Machine Dataset (SMD)}: This dataset consists of stacked trace data of server network resource utilization collected over five weeks by a large internet company. It contains data from 28 machines within a computing cluster, with each machine having 38 monitored metrics.\cite{su2019smd}

\textbf{Soil Moisture Active Passive (SMAP) Dataset}: Collected by NASA, this dataset comprises soil samples and telemetry information gathered by Mars probes for monitoring soil moisture sensor data. The dataset includes 55 entities, each with 25 monitored metrics.\cite{hundman2018smap}

\textbf{Mars Science Laboratory (MSL) Dataset}: Similar to the SMAP dataset but corresponds to sensor and actuator data from Mars probes themselves. It includes 27 entities, each with 55 monitored feature metrics.\cite{hundman2018msl}

\textbf{Pooled Server Metric (PSM) Dataset}: Collected by eBay, this dataset contains internal data from multiple eBay application server nodes. Specifically, the PSM dataset has 132,481 training data entries and 87,841 test data entries, where 13 weeks of data are used for training and 8 weeks for testing. There are 25 feature fields (from Feature 1 to Feature 25), along with a label field (0 indicates no anomaly, 1 indicates an anomaly; labels are provided only in the training set).\cite{abdulaal2021psm}

\textbf{Secure Water Treatment (SWaT) Dataset}: This dataset was collected from a real-world water treatment plant, containing 11 days of continuous operational data, including 7 days of normal operation and 4 days of anomalous operation. The dataset includes sensor values (such as water level, flow rates, etc.) and actuator operations (such as valve and pump actions).\cite{mathur2016swat}

\begin{figure*}[!htb]
  \centering
  \begin{subfigure}[b]{0.33\textwidth}
      \includegraphics[width=\textwidth]{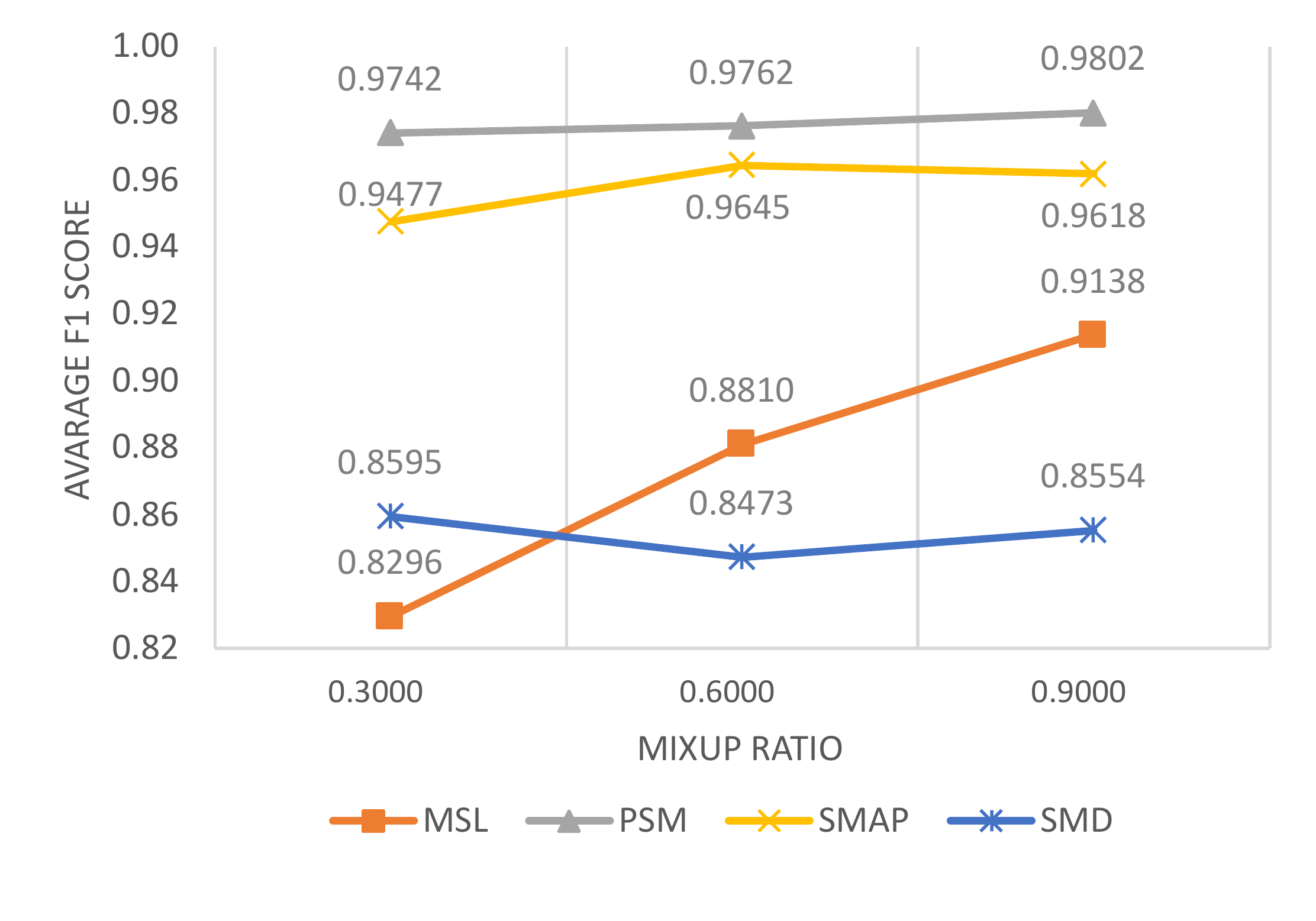}
      \caption{Average F1 for $\alpha$}
      \label{fig:gridsubfig1}
  \end{subfigure}
  \hfill
  \begin{subfigure}[b]{0.32\textwidth}
      \includegraphics[width=\textwidth]{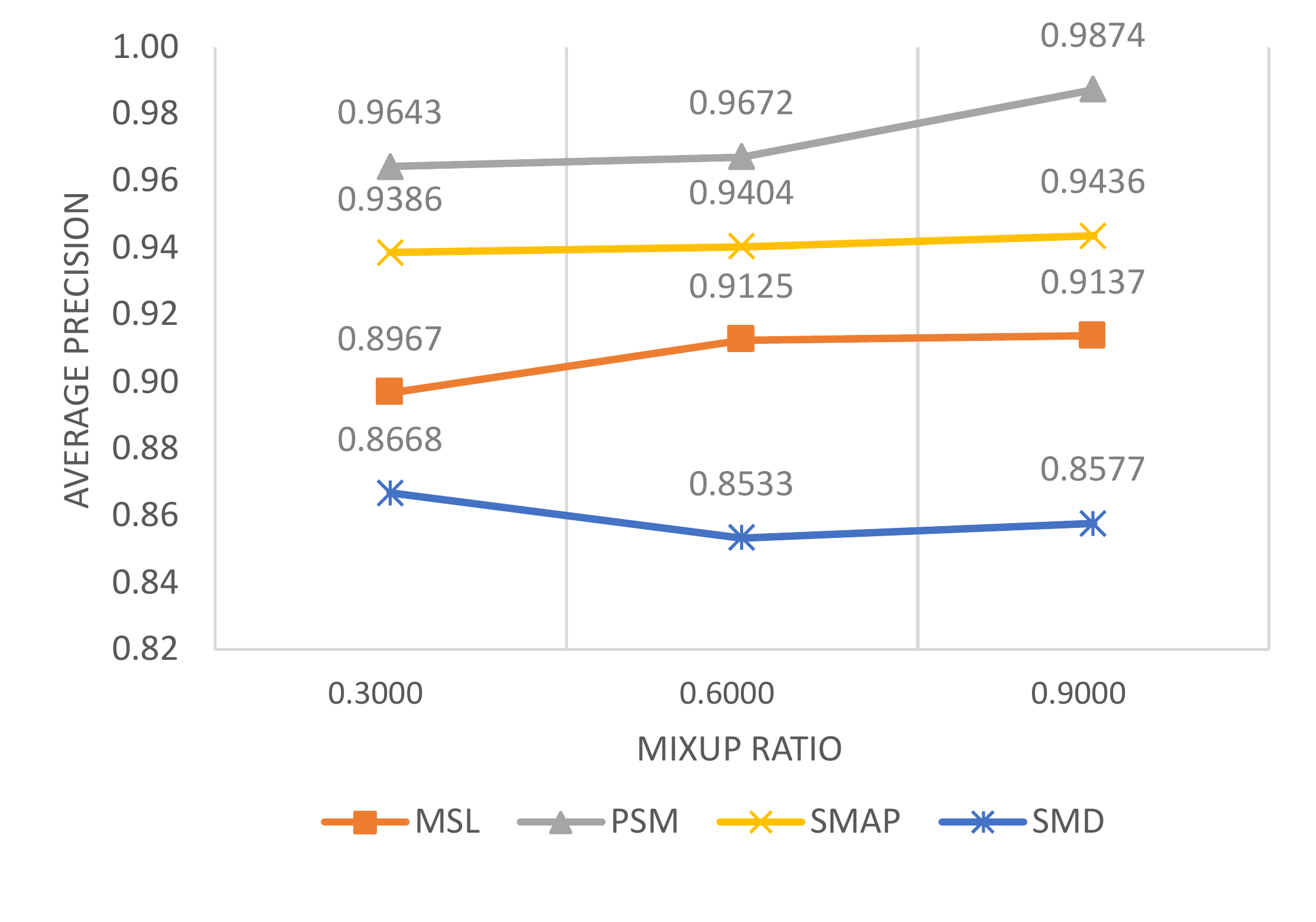}
      \caption{Average Precision for $\alpha$}
      \label{fig:gridsubfig3}
  \end{subfigure}
  \vspace{1em}
  \begin{subfigure}[b]{0.32\textwidth}
      \includegraphics[width=\textwidth]{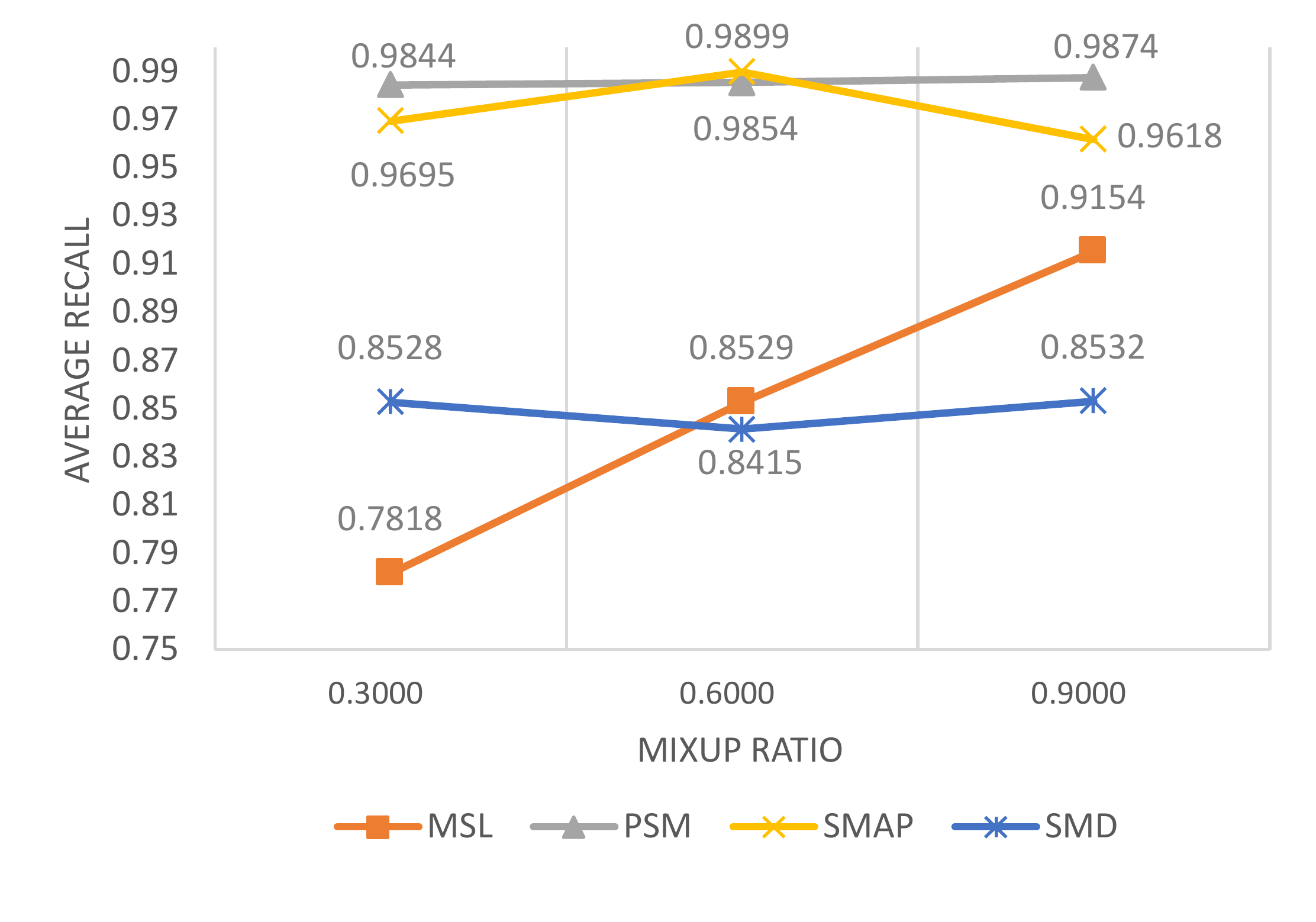}
      \caption{Average Recall for $\alpha$}
      \label{fig:gridsubfig4}
  \end{subfigure}
  \hfill
  \begin{subfigure}[b]{0.32\textwidth}
      \includegraphics[width=\textwidth]{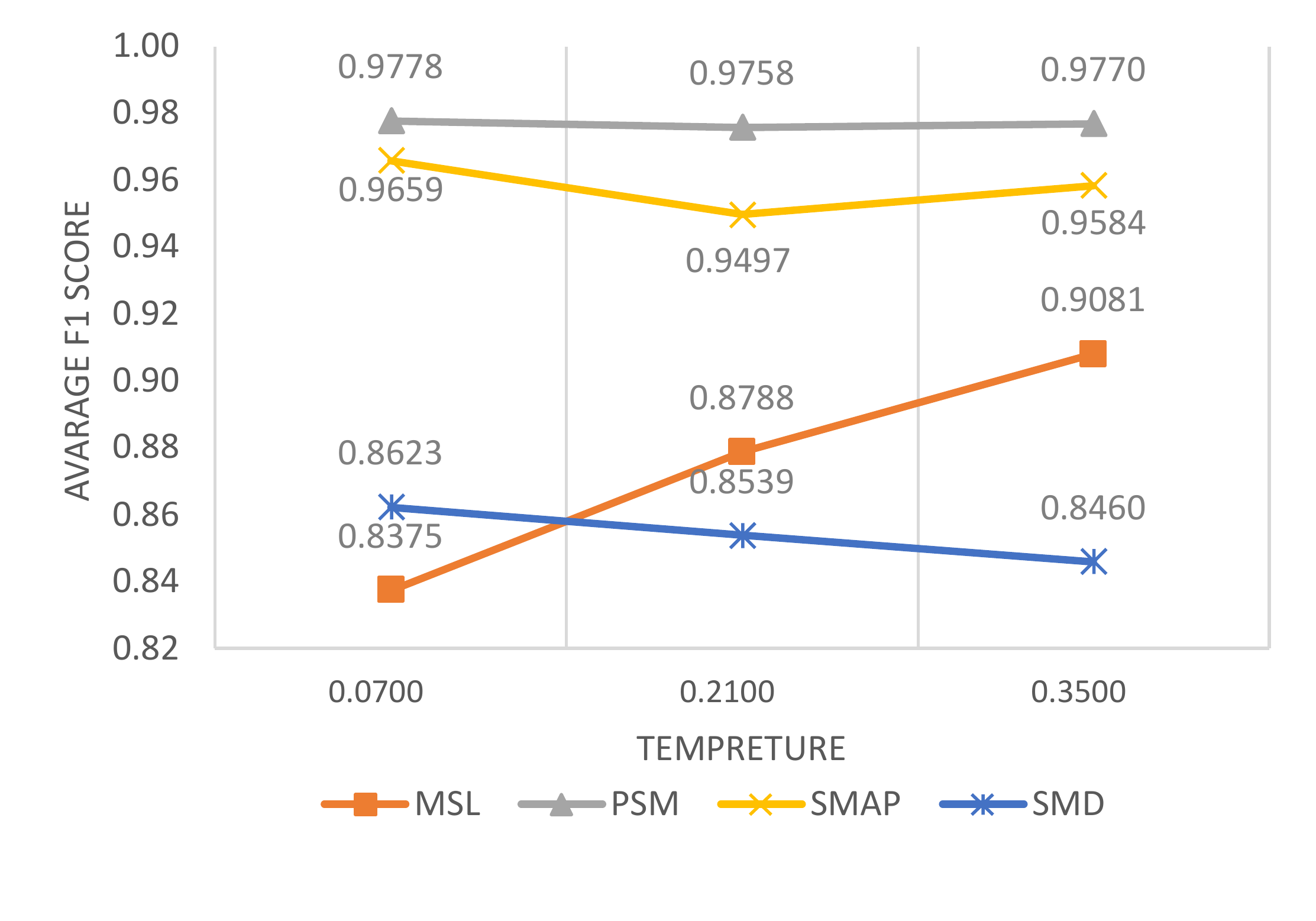}
      \caption{Average F1 for $\tau$}
      \label{fig:gridsubfig2}
  \end{subfigure}
  \hfill
  \begin{subfigure}[b]{0.32\textwidth}
      \includegraphics[width=\textwidth]{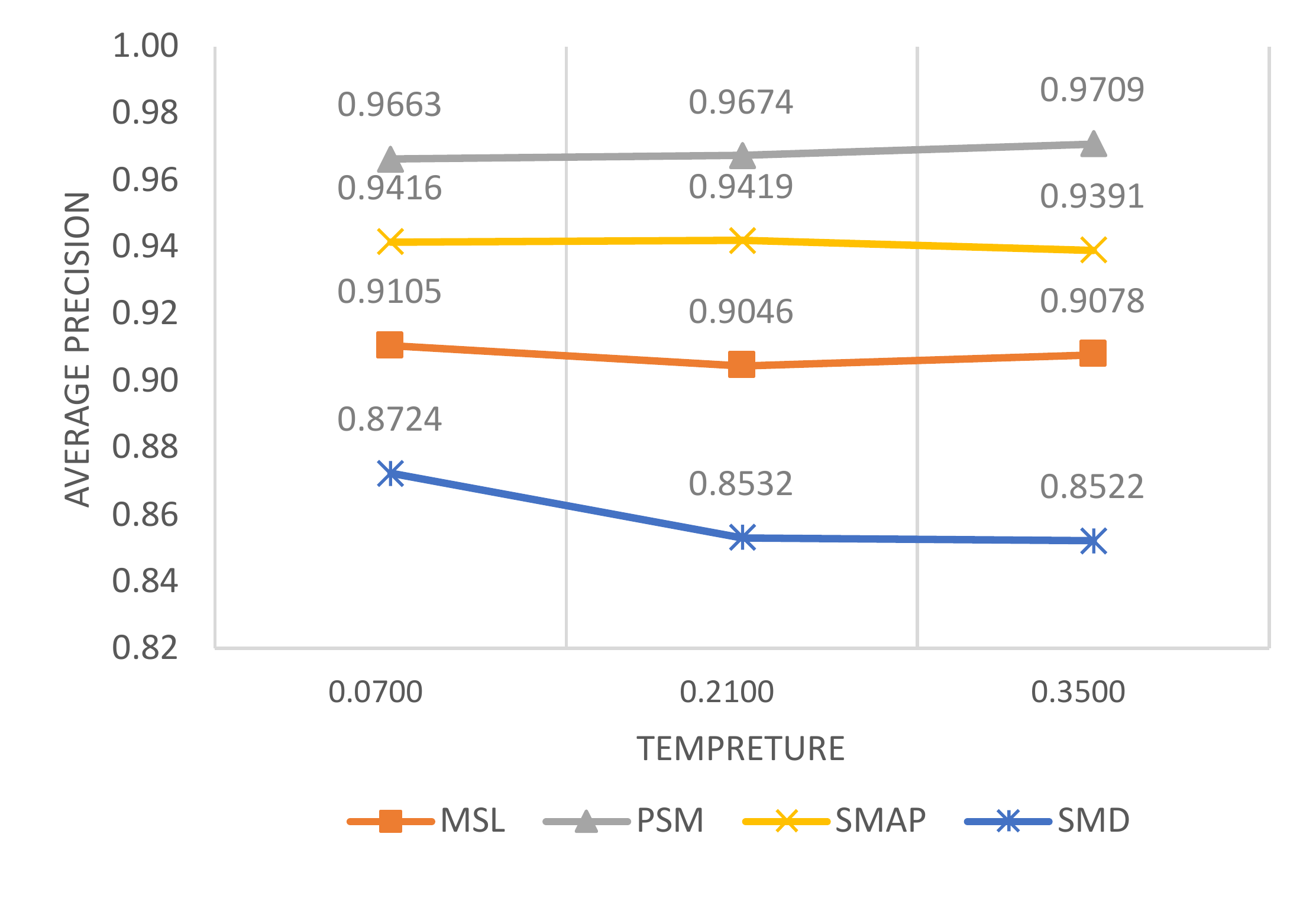}
      \caption{Average Precision for $\tau$}
      \label{fig:gridsubfig5}
  \end{subfigure}
  \hfill
  \begin{subfigure}[b]{0.32\textwidth}
      \includegraphics[width=\textwidth]{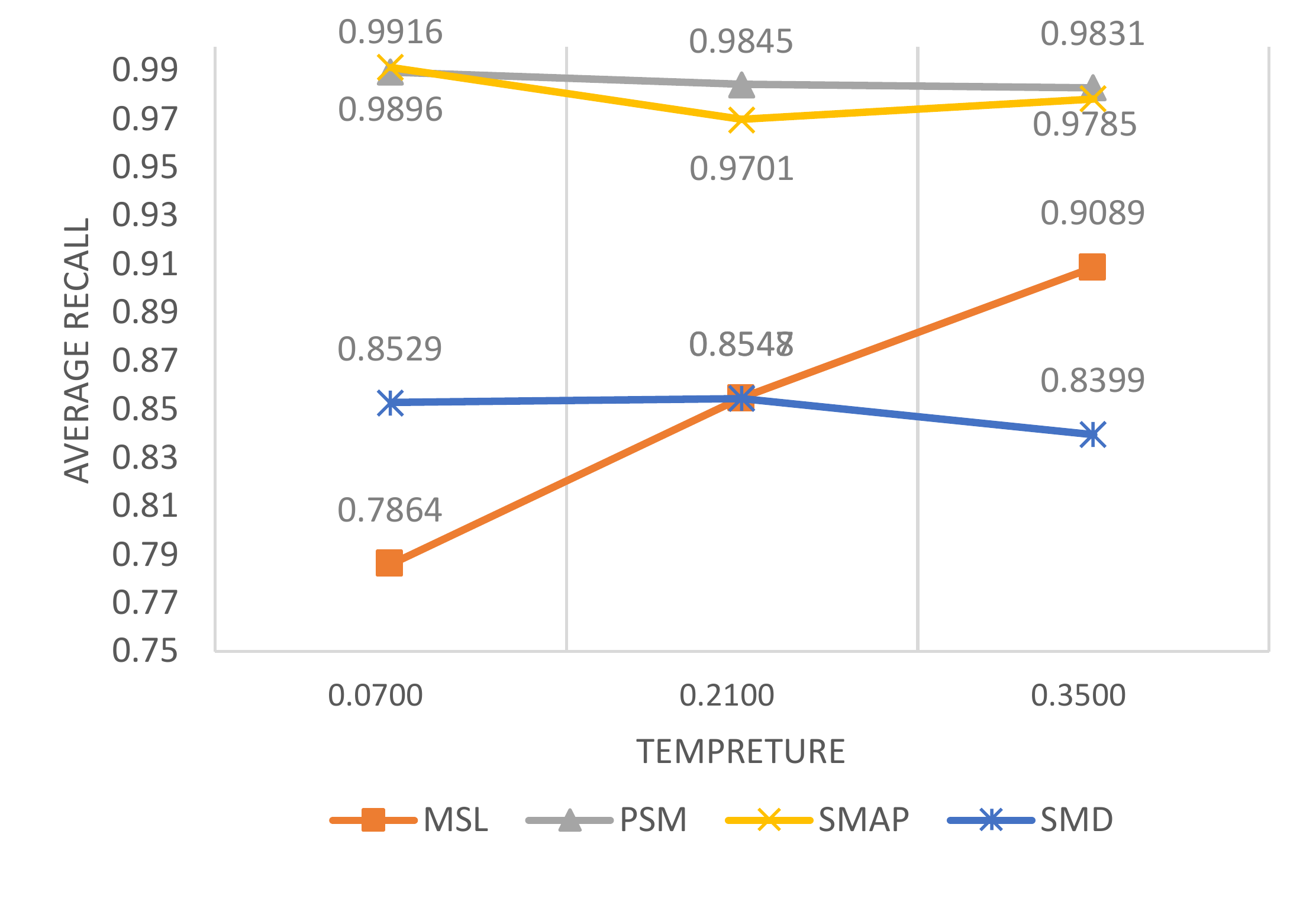}
      \caption{Average Recall for $\tau$}
      \label{fig:gridsubfig6}
  \end{subfigure}
  \caption{Hyperparameter Search Results}
  \label{fig:gridsearch}
\end{figure*}

\subsection{Performance Results}
\label{sec:amadexperimentset}

Here we present a detailed performance comparison of the proposed AMAD model with state-of-the-art baselines across five anomaly detection datasets (MSL, SWaT, PSM, SMAP, and SMD). \autoref{tab:amad_results} presents the experimental results of our model and baseline methods. $ar$ denotes the prior anomaly ratio used for percentile calculation. We report P (Precision), R (Recall), and F1 (F1-score), with F1 being the harmonic mean of P and R. The best results are in bold, and the second best are underlined.

We compared our model with 11 baseline methods: MAD-GAN, THOC, InterFusion, BeatGAN, DAGMM, MTAD-GAT, LSTM-AD, OmniAnomaly, TranAD, AnomalyTransformer, and ImDiffusion \cite{li2019madganma,THOC,InterFusion,zhou2019beatganar,zhao2020mtadgat,malhotra2015lstmad,su2019ominianomaly,tuli2022tranad,xu2021anomalytrans,chen2023imdiffusion}. Among these, AnomalyTrans and ImDiffusion are SOTA methods. It is worth to note that the post adjustment method proposed by early works adopted as a regular step before measurements \cite{msadren2019time,usadkdd20,tuli2022tranad}, for a comparable propose, we decided not to break this tradition, with the same adjustment method as \cite{xu2021anomalytrans}.

AMAD demonstrates superior or competitive performance across most datasets, particularly excelling in PSM and SMAP, where it achieves the highest F1-scores (0.9817 and 0.9683, respectively). Notably, AMAD outperforms all baselines in PSM by achieving the highest Recall (0.9910) and F1, indicating robust anomaly detection without compromising precision (P=0.9726). In SMAP, AMAD's Recall (0.9948) and F1 (0.9683) are unmatched, highlighting its ability to detect nearly all anomalies while maintaining high precision. However, AMAD exhibits moderate performance on SMD (F1=0.8748), lagging behind ImDiffusion (0.9488), suggesting potential limitations in handling imbalanced datasets (ar=0.5).

Separately analyzing the results for each dataset, we observe the following trends:
 
In MSL,
AMAD achieves the second-highest Precision (0.9190) and the highest Recall (0.9569), yielding an F1 of 0.9375. While AnomalyTransformer leads in Precision (0.9209), its Recall (0.9515) is slightly lower. BeatGAN shows a trade-off between high Recall (0.8746) and low Precision (0.6401), indicating over-detection. AMAD effectively balances precision and recall, making it suitable for high-anomaly-density datasets like MSL.

In SWaT,
AMAD achieves the second-highest Precision (0.9844) but underperforms in Recall (0.7134), resulting in an F1 of 0.8273. LSTM-AD leads in Recall (0.9925) but struggles with Precision (0.6737), suggesting overfitting. AnomalyTransformer attains the highest Recall (0.9673) but lower Precision (0.9155). AMAD's prioritization of precision over recall may be beneficial for avoiding false positives in critical industrial systems like SWaT (a water treatment dataset).

In PSM,
AMAD dominates with the highest Precision (0.9726), Recall (0.9910), and F1 (0.9817), outperforming all baselines. InterFusion and TranAD trail far behind in Recall (0.8345 and 0.8951, respectively). AMAD's architecture, likely incorporating advanced temporal modeling, excels in capturing complex patterns in PSM, a process system dataset.

In SMAP,
AMAD achieves the highest Recall (0.9948) and F1 (0.9683), outperforming AnomalyTransformer (F1=0.9636). ImDiffusion struggles with Recall (0.9618) compared to AMAD. AMAD's design effectively handles spatial-temporal anomalies in soil moisture sensor data, likely due to its ability to model long-range dependencies.

In SMD, 
MTAD-GAT leads in Precision (0.9718) but has weak Recall (0.5259) and F1 (0.6824). ImDiffusion achieves the highest F1 (0.9488) due to strong Recall (0.9509) and Precision (0.9520). AMAD underperforms with F1=0.8748, suggesting challenges in imbalanced datasets. This highlights the need for adaptive strategies in low-anomaly-ratio scenarios.

The results show that the AMAD model we proposed achieves excellent F1 scores across multiple datasets. It attains the best Recall and F1 values on the MSL, PSM, and SMAP datasets, delivering SOTA performance. The model also shows strong competitiveness on the other two datasets. This indicates that our model achieves good performance across multiple datasets and has strong capabilities in time series anomaly representation and detection.

We summarize the key strengths of our model performance as follows:

\textbf{High Recall in Critical Datasets}: Near-perfect Recall in PSM and SMAP, critical for industrial/environmental monitoring.

\textbf{Balanced Precision-Recall Trade-off}: Outperforms most baselines in F1 across four datasets.

\textbf{Complex Scenario Adaptation}: Excels in datasets requiring temporal/spatial reasoning (PSM, SMAP), likely due to hybrid feature encoding or attention mechanisms.

\subsection{Hyperparameter Search for $\alpha$ and $\tau$}
We conducted an extensive hyperparameter search on four datasets: MSL, PSM, SMAP, and SMD, using the grid method to analyze the impact of two hyperparameters in our model's attention fusion Mixup module: the mixing coefficient $\alpha \in \{0.3, 0.6, 0.9\}$ and the temperature parameter $\tau \in \{0.07, 0.21, 0.35\}$. The results, averaged over the edges, are shown in \autoref{fig:gridsearch}.

From the results, we empirically conclude that smaller $\alpha$ values should be paired with smaller $\tau$ values, and larger $\alpha$ with larger $\tau$. This may be because a higher proportion of AutoMask attention requires a higher $\tau$ to enhance contrastive learning sensitivity. Additionally, we found that larger $\alpha$ values yield better performance, indicating that the AutoMask block effectively learns more information than Self Attention.

\subsection{Ablation Study Results}

\begin{table*}[!htbp]
  \centering
  \caption{Ablation Study Results}
  \scalebox{0.92}{
    \begin{tabular}{c|c|c|c|c|c|c|c|c|c|c}
    \toprule
    \multicolumn{4}{c|}{\multirow{2}[2]{*}{Model Resection}} & \multicolumn{3}{c|}{\multirow{2}[2]{*}{SMAP}} & \multicolumn{3}{c|}{\multirow{2}[2]{*}{SMD}} & \multirow{3}[4]{*}{Avg F1} \\
    \multicolumn{4}{c|}{} & \multicolumn{3}{c|}{} & \multicolumn{3}{c|}{} &  \\
\cmidrule{1-10}    Min Strategy & Max Strategy & Contrastive Strategy & AutoMask Module & P   & R   & F1  & P   & R   & F1  &  \\
    \midrule
    w/o & w/o & w/o & w/o & 0.7508 & 0.6938 & 0.7212 & 0.8225 & 0.7356 & 0.7766 & 0.7489 \\
    w/o & w/o & \checkmark & \checkmark & 0.9375 & 0.9874 & 0.9618 & 0.8738 & 0.8390 & 0.8560 & 0.9089 \\
    \checkmark & \checkmark & w/o & \checkmark & 0.9359 & 0.9948 & 0.9644 & 0.8739 & 0.8446 & 0.8590 & 0.9117 \\
    w/o & \checkmark & w/o & \checkmark & 0.8891 & 0.9451 & 0.9162 & 0.8708 & 0.8043 & 0.8363 & 0.8763 \\
    w/o & w/o & w/o & \checkmark & 0.8913 & 0.9406 & 0.9153 & 0.8273 & 0.7641 & 0.7944 & 0.8548 \\
    \midrule
    \checkmark & \checkmark & \checkmark & \checkmark & \textbf{0.9432} & \textbf{0.9948} & \textbf{0.9683} & \textbf{0.9043} & \textbf{0.8471} & \textbf{0.8748} & \textbf{0.9216} \\
    \bottomrule
    \end{tabular}%
  }
  \label{tab:ablation_results}%
\end{table*}%

We conducted comprehensive ablation studies on the \textbf{SMAP} and \textbf{SMD} datasets, covering four components of our model: the Min/Max strategy, contrastive strategy, AutoMask module, and an additional component. We carried out five sets of experiments, using “w/o” to denote the removal of a specific component and \checkmark to indicate its retention. Removing all four modules means resorting only to the classic Transformer model, while eliminating all proxy task modules implies training solely with the reconstruction loss.

\begin{figure}[!h]
    \centering
    \includegraphics[width=.5\textwidth]{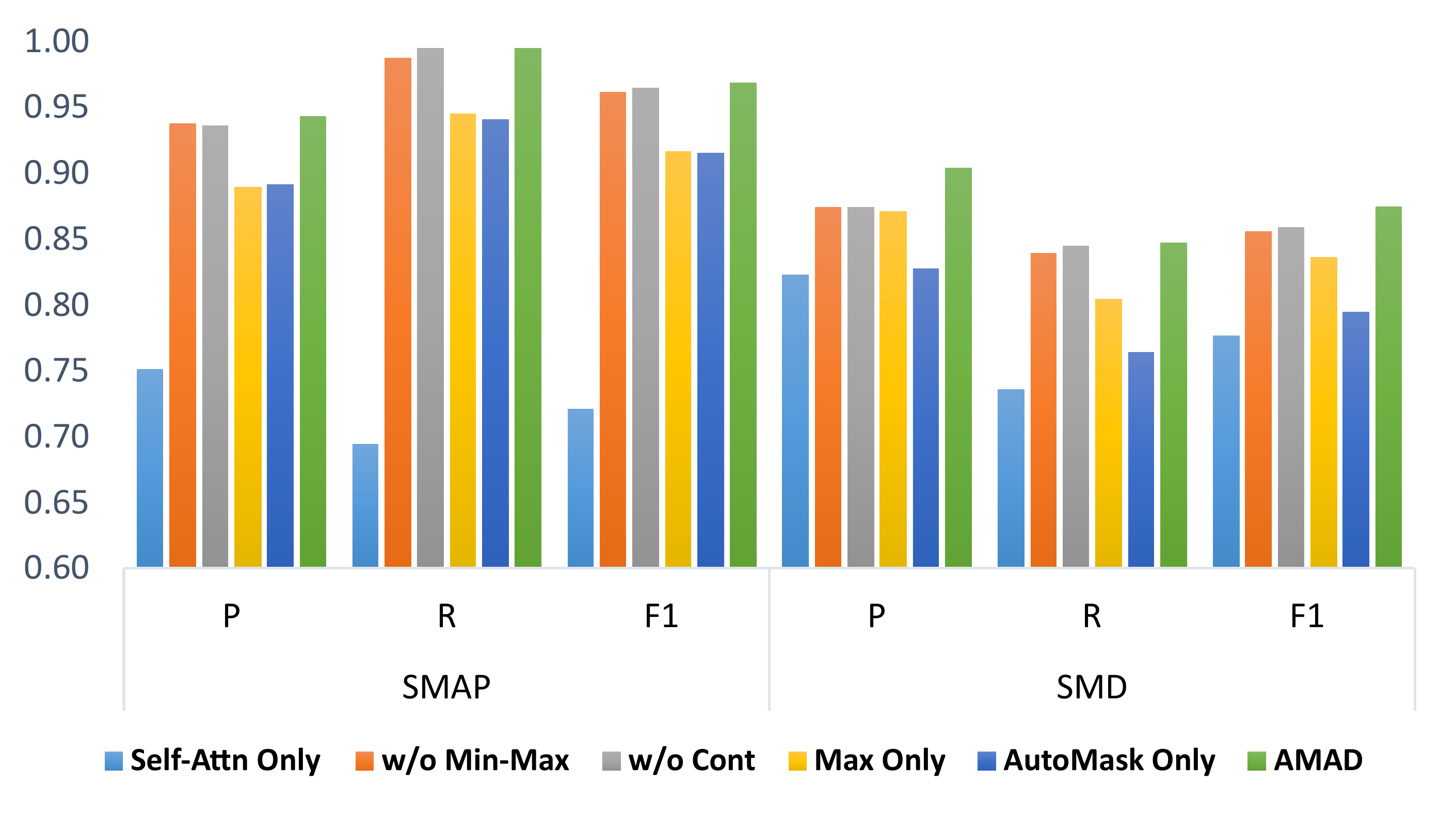}
    \caption{Ablation Study Results of the AMAD Model}
    \label{fig:amad_ablation_results}
\end{figure}

As shown in \autoref{tab:ablation_results}, we progressively removed the Min/Max strategy, contrastive strategy, and AutoMask module from our model. Results indicate that, compared to using only the Transformer model, AMAD achieves over 20\% improvement. 

Removing Min/Max strategy causes around 5\% performance drop on two datasets. This highlights the significant impact of the Min/Max strategy and AutoMask module, and shows that their combined effect is essential for optimal performance.

The AutoMask module has the most significant impact on SMD (highly imbalanced data), while the Min/Max strategy is critical for SMAP (high-dimensional data).
The contrastive strategy consistently improves performance across both datasets, highlighting its role in enhancing anomaly discrimination.

The ablation study confirms that each component of AMAD contributes uniquely to its performance. The Min/Max strategy and AutoMask module are indispensable for handling outlier detection and overfitting, respectively, while the contrastive strategy and additional proxy tasks provide complementary supervision. The results validate the necessity of the full architecture for achieving state-of-the-art performance on diverse anomaly detection tasks. Future work may explore lightweight variants of these components to improve efficiency while retaining effectiveness.

The results also highlight that the synergistic combination of components (e.g., contrastive learning with AutoMask) is critical for robust anomaly detection. For instance, the AutoMask module’s dynamic feature masking prevents overfitting to noisy features, while the Min/Max strategy ensures robustness to extreme values. These findings align with prior work on multi-task learning and self-supervised methods, which emphasize the importance of complementary objectives for complex tasks.

\section{Conclusion and Future Work}

AMAD draws inspiration from the Fourier Transformation, which decomposes functions into a spectrum of periodic components. Analogously, AMAD introduces an AutoMask attention mechanism that acts as a spectral decomposition for time series data. By embedding learnable rotary positional encodings, AMAD generalizes beyond fixed Gaussian kernels to approximate arbitrary correlation functions, enabling the model to capture both local and global sequence features effectively.

The AutoMask mechanism dynamically modulates sequence representations across multiple scales, akin to a spectral basis in Fourier analysis. This allows AMAD to model complex temporal dependencies and anomaly correlations that are otherwise challenging to capture. The model employs a Max-Min training strategy to balance local and global feature learning, ensuring robust anomaly detection without descending into trivial solutions. Additionally, the attention fusion module integrates multi-scale features through a softmax mixup operation, enhancing the model's adaptability to diverse anomaly patterns.

In summary, AMAD represents a novel approach to unsupervised multivariate time series anomaly detection, leveraging spectral decomposition principles to achieve competitive performance, demonstrating its robustness in capturing multi-scale anomaly correlations. However, challenges remain in handling imbalanced datasets and optimizing computational efficiency. Future work could explore adaptive mechanisms for varying anomaly ratios and incorporate advanced geometric representations, such as hyperbolic embeddings, to further enhance the model's capability in representing complex temporal structures.

\begin{acks}
%  This work was supported by the [...] Research Fund of [...] (Number [...]). Additional funding was provided by [...] and [...]. We also thank [...] for contributing [...].
Thanks to Joanny Huang for her help with the English grammar, making this paper much clearer.
\end{acks}

\bibliographystyle{ACM-Reference-Format}
\bibliography{cited}

\end{document}